\documentclass[10pt,twocolumn,letterpaper]{article}

\usepackage{cvpr}              
\usepackage{times}

\usepackage{multicol}
\usepackage{multirow}
\usepackage{makecell}

\usepackage{colortbl}
\usepackage{pifont}
\usepackage{comment}
\usepackage[T1]{fontenc}

\definecolor{cvprblue}{rgb}{0.21,0.49,0.74}
\definecolor{lavender}{rgb}{0.8, 0.6, 1.0}

\usepackage[pagebackref,breaklinks,colorlinks,allcolors=cvprblue, pdfa=true]{hyperref}

\def\paperID{01106} 
\def\confName{CVPR}
\def\confYear{2025}

\title{\includegraphics[scale=0.03]{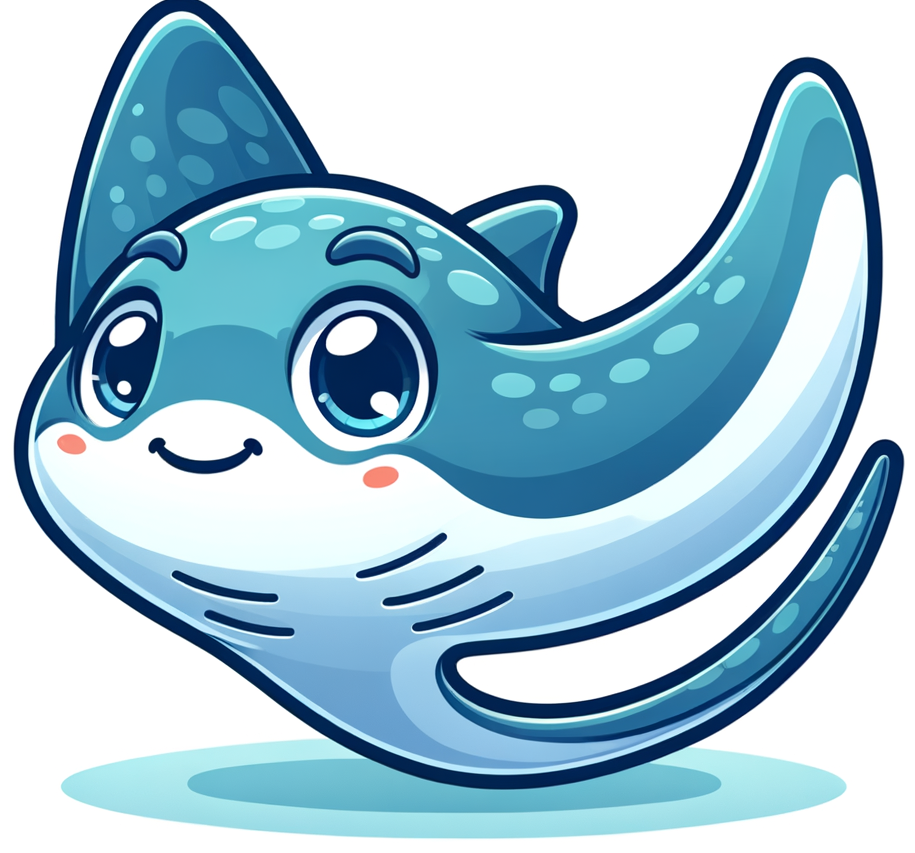} MANTA: Diffusion Mamba for Efficient and Effective Stochastic Long-Term Dense Action Anticipation}
\author{Olga Zatsarynna\textsuperscript{1,4}, Emad Bahrami\textsuperscript{1,4}, Yazan Abu Farha\textsuperscript{2}, Gianpiero Francesca\textsuperscript{3}, Juergen Gall\textsuperscript{1,4} \\
\small{\textsuperscript{1}University of Bonn, \textsuperscript{2}Birzeit University, \textsuperscript{3}Toyota Motor Europe,  \small{\textsuperscript{4}Lamarr Institute for Machine Learning and Artificial Intelligence}}\\
}

\begin{document}
\maketitle

\begin{abstract}
Long-term dense action anticipation is very challenging since it requires predicting actions and their durations several minutes into the future based on provided video observations. To model the uncertainty of future outcomes, stochastic models predict several potential future action sequences for the same observation. Recent work has further proposed to incorporate uncertainty modelling for observed frames by simultaneously predicting per-frame past and future actions in a unified manner. While such joint modelling of actions is beneficial, it requires long-range temporal capabilities to connect events across distant past and future time points. However, the previous work struggles to achieve such a long-range understanding due to its limited and/or sparse receptive field.
To alleviate this issue, we propose a novel MANTA (MAmba for ANTicipation) network. Our model enables effective long-term temporal modelling even for very long sequences while maintaining linear complexity in sequence length. We demonstrate that our approach achieves state-of-the-art results on three datasets—Breakfast, 50Salads, and Assembly101—while also significantly improving computational and memory efficiency. Our code is available at \href{https://github.com/olga-zats/DIFF_MANTA}{https://github.com/olga-zats/DIFF\_MANTA}.
\end{abstract}

\footnotetext[1]{Corresponding author: zatsarynna@iai.uni-bonn.de}

\vspace{-0.2cm}
\section{Introduction}
In this work, we address the task of stochastic long-term dense anticipation. This task focuses on forecasting actions and their durations over minutes-long future horizons. The length of the anticipation horizon and the need to predict the duration of future actions make this task particularly challenging. However, solving it is essential for many real-world applications, including autonomous driving and assembly line automation.  

In our work, we focus on stochastic anticipation, where multiple predictions for a single video observation are made to account for the uncertainty of future forecasting. Early works in this area~\cite{farha2019uaaa, zhao2020async} predicted future action sequences based on a fixed set of pre-classified or ground-truth observed actions. Recently, GTDA~\cite{zatsarynna2024gtd} highlighted the limitations of this approach and proposed to model framewise past and future actions jointly with a single diffusion model, conditioned on the observed frames padded with zeros in place of future entries. To account for the differences between the observed and masked parts of the sequence, GTDA introduced a GTAN diffusion generator - a multi-stage architecture of dilated temporal convolutional layers with data-dependent temporal gates.

\begin{figure}
    \centering
    \includegraphics[width=1.0\columnwidth]{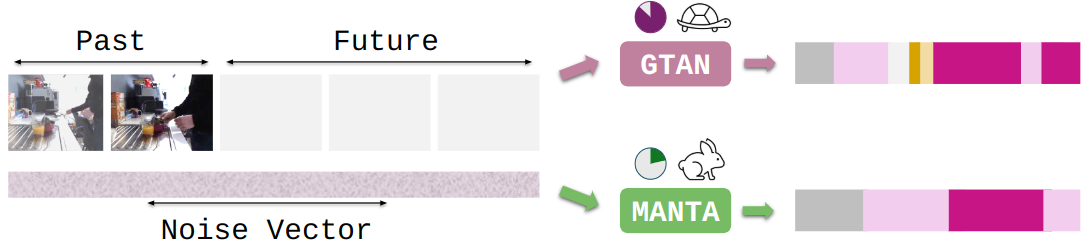}
    \vspace{-0.4cm}
    \caption{We propose a novel MANTA diffusion generator that allows for more efficient and effective stochastic long-term anticipation compared to the previous work.}
    \vspace{-0.5cm}
    \label{fig:teaser}
\end{figure}

Although it achieves strong performance on the task of stochastic long-term anticipation, the proposed GTAN generator has certain limitations. Firstly, in the early layers of each stage, convolutional kernels have local support due to the small dilation factor. However, given the input structure, where observed frames are followed by masking values as depicted in Fig.~\ref{fig:teaser}, the limited receptive field prevents padded values from accessing observed context and thus updating future frame values. While the multi-stage architecture partially mitigates this, it introduces additional parameters and increases computation. Moreover, while in the later layers, the dilation rate increases the receptive field, it remains sparse due to the small kernel size, with each sequence element accessing only a limited set of other elements, making updates highly restrictive. These limitations are particularly pronounced for longer videos with extended anticipation horizons, where the future-padded part of the input sequence becomes challenging to process.

To address these limitations, we propose a novel MANTA (MAmba for ANTicipation) generator network, which incorporates the recently proposed Mamba layer~\cite{gu2023mamba} as its temporal processing block. Our resulting model maintains a global receptive field while keeping linear complexity relative to the sequence length, enabling early propagation of observed information across the entire masked area of the input sequence. This eliminates the need for a multi-stage architecture and reduces computational and memory costs. Additionally, our model maintains the ability to adaptively handle the two distinct parts of the input sequence - past observed frames and future masked frames - due to the inherent data-dependent gating mechanism.

We evaluate our proposed model on three challenging datasets: Breakfast~\cite{Kuehne12}, 50Salads~\cite{Stein2013CombiningEA}, and Assembly101~\cite{sener2022assembly101}. Our results demonstrate state-of-the-art performance across all three of them. Furthermore, we show, that our model is significantly faster than the previous best-performing GTDA~\cite{zatsarynna2024gtd} model, achieving inference and training time speed-up of up to $65.3\times$ and $6.6\times$, respectively.

\section{Related Work}
\label{sec:related_work}

\textbf{Action Anticipation in Videos.}
Action anticipation has been explored in various works across different settings. Although the overarching goal remains to predict future actions based on observed video data, the key distinction lies in the temporal horizon of anticipation. On one hand, short-term anticipation approaches~\cite{girdhar2021anticipative, zatsarynna_2022_gcpr, zatsarynna2021MMTCN, zatsarynna2023goal, furnari2020rulstm, Zhong2022AnticipativeFF, Liu_2020_ECCV, zhao2022testra, sener2020temporal, hongji2024unc, debatiya2024inter, harshayu2023latency} focus on predicting a single action a few seconds before it occurs. Long-term anticipation methods, on the other hand, aim to predict multiple future actions, resulting in forecasting horizons that span one or several minutes. Here, we focus on the latter setting.

In the context of long-term anticipation, several distinct settings exist that differ in their final prediction objectives. The focus of this work is \textit{dense long-term anticipation}.
Unlike different research directions that frame anticipation as ordered~\cite{Grauman2021Ego4DAT, Mascaro_2023_WACV, Ashutosh_2023_CVPR, Das2022VideoC, Zhao2023AntGPTCL, mittal2024can} or undordered~\cite{nawhal2022anticipatr, ego-topo, zhong2023diffant, Zhao2023AntGPTCL} duration-agnostic transcript prediction problem, dense anticipation requires future actions to be predicted for a predefined number of future frames. This involves the estimation of both the order of actions and their durations. Dense long-term anticipation is further subdivided into deterministic and stochastic anticipation. Deterministic models~\cite{gong2022future, farha2020gcpr, Farha_2018_CVPR, zatsarynna2024gtd, zhong2023diffant, sener2020temporal, Ke_2019_CVPR} produce a single prediction for each video observation, while stochastic methods~\cite{farha2019uaaa, zatsarynna2024gtd, zhong2023diffant, zhao2020async} generate several plausible samples for each observation. In this work, we propose a stochastic long-term dense anticipation approach.

Farha~\etal~\cite{farha2019uaaa} introduced the first stochastic approach for dense anticipation by extending a deterministic RNN-based method~\cite{Farha_2018_CVPR} to predict multiple samples. Zhao~\etal~\cite{zhao2020async} enhanced the plausibility and diversity of these predictions through adversarial learning. Instead of relying on pre-classified or ground-truth actions like previous approaches, recent methods~\cite{zatsarynna2024gtd, zhong2023diffant} perform past classification and future forecasting simultaneously. Zhong~\etal~\cite{zhong2023diffant} expanded the FUTR~\cite{gong2022future} decoder with diffusion~\cite{sohl2015NonEquTherm, song2019generative, ho2020denoisingDiff, song2021scorebased, ramesh2022hierarchical} to model uncertainty in future actions. Zatsarynna~\etal~\cite{zatsarynna2024gtd} proposed a gated temporal diffusion network to simultaneously model uncertainty in past and future actions, outperforming previous methods.

\textbf{State-Space Models.}
Recent developments in sequence modeling have introduced efficient alternatives to attention-based architectures, such as State Space Models~\cite{gu2022efficientlyS4, smith2023simplifiedS6, NEURIPS2021_05546b0e}, which offer linear scaling with sequence length. Building on these approaches, Mamba~\cite{gu2023mamba} enhances the framework by introducing selective state space modeling with an input-dependent selection mechanism. Given Mamba's potential for efficient modeling of long sequences, it has been adapted for tasks such as image classification~\cite{liu2024vmamba, zhu2024visionMamba, li2025mambaND}, video understanding~\cite{li2025videomamba, wang2023selectiveVideo, li2025mambaND}, medical imaging~\cite{ma2024uMamba, ruan2024VmUnet, xing2024segmamba}, image restoration~\cite{guo2025mambair}, motion generation~\cite{zhang2025motion} and point cloud analysis~\cite{liang2024pointmamba}. Architectures based on state space models such as Mamba have recently been used as backbones for diffusion models in image and video generation~\cite{hu2024zigma, mo2024scaling, yan2024diffusion}. 
In this work, distinct from previous studies, we focus on designing a Mamba-based model for both efficient and effective stochastic long-term action anticipation.

\section{Method}
\label{sec:motivation}
To effectively handle the inherent uncertainties in action forecasting, we explore stochastic long-term anticipation, where models predict multiple possible future outcomes based on a single set of observed frames. Given the observed frames, these approaches learn to sample from the probability distribution of future action variables. Formally: 
\begin{align}
    \tilde{\mathcal{Y}}^{P+1:P+F} & \sim P\big( \mathcal{Y}^{P+1:P+F} \mid g(x_1):g(x_P) \big),
\end{align}
where \( g \) is a feature extractor, \( x_i \) is the $i^{\text{th}}$ video frame, \( \mathcal{Y}^{i} \) is the variable corresponding to frame $i$, and \( \tilde{\mathcal{Y}}^{i} \) is a sample from $\mathcal{Y}^{i}$; \( P \) and \( F \) represent the number of past and future frames, respectively. 
In~\cite{zatsarynna2024gtd}, the uncertainty of actions in both observed and future frames is jointly modelled by a more general distribution - the joint observation-conditioned probability of both observed and future actions:
\begin{align}
\label{eq:formulation}
\tilde{\mathcal{Y}}^{1:P+F} & \sim P\big( \mathcal{Y}^{1:P+F} \mid g(x_1):g(x_P) \big),
\end{align}
and a common diffusion model predicts actions in the future and past frames in a consistent manner.
Furthermore, to take into account distinctions between these two types of variables, \cite{zatsarynna2024gtd} proposed a GTAN diffusion generator network, that employs dilated gated temporal convolutional layers to process the sequence with data-dependent gates. The use of dilated convolutions for temporal processing, however, poses several limitations. Specifically, the exponentially increasing dilation rates ($2^{l-1}$) of the convolutional kernels result in limited receptive fields in the early layers and sparse receptive fields lacking local context in the later layers. This leads to two issues: the limited receptive field of early layers prevents observed information from reaching future masked entries, while the large, sparse receptive fields of the later layers restrict updates and lack local awareness. Although the multi-stage design intends to address these issues, it only partially mitigates them, while adding to computational and memory costs.

In this work, we address the limitations mentioned above. While we follow Zatsarynna~\etal~\cite{zatsarynna2024gtd} in harnessing the unified diffusion process for observed and future action modelling, we introduce a novel MANTA generator model. By utilizing Mamba~\cite{gu2023mamba}-based layers, our model sustains a global receptive field across the entire network while preserving local awareness. Additionally, relying on the inherent data-dependent gating, our model processes the input sequence adaptively, considering the difference between its two components - observed and future actions.
In the following sections, we present the details of our proposed model. In Sec.~\ref{sec:diffusion}, we describe how the diffusion process is formulated for addressing the task of stochastic long-term dense anticipation. Then, in Sec.~\ref{sec:model}, we present the details of our proposed MANTA model.

\subsection{Diffusion for Long-Term Dense Anticipation}
\label{sec:diffusion}

\textbf{General Diffusion}. Diffusion models are generative models that learn to produce samples from data distributions by reconstructing progressively noisier versions of data points. To achieve this, two Markov chain processes are defined: the \textit{forward} and the \textit{reverse}. The forward Markov process $ q(\mathcal{Y}_{t} | \mathcal{Y}_{t-1}) $ defines the transition from the data distribution to Gaussian noise, using a predefined noising schedule. In contrast, the reverse Markov process $ p_\theta(\mathcal{Y}_{t-1} | \mathcal{Y}_t)$ models the inverse transition, from random noise back to the data distribution.
The reverse process is parameterized by a diffusion generator network \( \mathcal{G}_{\theta} \) that is trained to reconstruct data points corrupted by the forward process: ${\tilde{\mathcal{Y}}_{0,t}} = \mathcal{G}_\theta(\mathcal{Y}_t, t)$. 
Here,  $\mathcal{Y}_t$ is the corrupted data sample and $\tilde{\mathcal{Y}}_{0, t}$ is the reconstructed data point. 

\textbf{Anticipation Diffusion.} We adopt the formulation of the diffusion anticipation model proposed in~\cite{zatsarynna2024gtd}. 
Following Eq.~\eqref{eq:formulation}, anticipation diffusion reconstructs one-hot encoded per-frame past and future ground-truth actions conditioned on the padded visual features of observed frames. More specifically, the diffusion latent variables \( \mathcal{Y}_t \) are introduced to model per-frame actions, with a total of \( (P+F) \) variables. To condition anticipation on past video frames, latent variables $\mathcal{Y}_t$ are concatenated with a vector $\mathcal{X}$, constructed by extending observed visual features with zeros in place of future action variables to account for the corresponding unobservable visual features:
\begin{align}
\label{eq:padding}
& \mathcal{X} = \{ g(x_1), \ldots, g(x_P), \underset{F}{0, \ldots, 0}\},
\end{align}
In this way, the future reconstruction is computed as $\mathcal{G}_\theta(\mathcal{Y}_t, t, \mathcal{X})$. Next, we provide a detailed description of our proposed MANTA generator $\mathcal{G}_\theta$.

\begin{figure*}[t!]
    \centering
    \includegraphics[width=1.\textwidth]{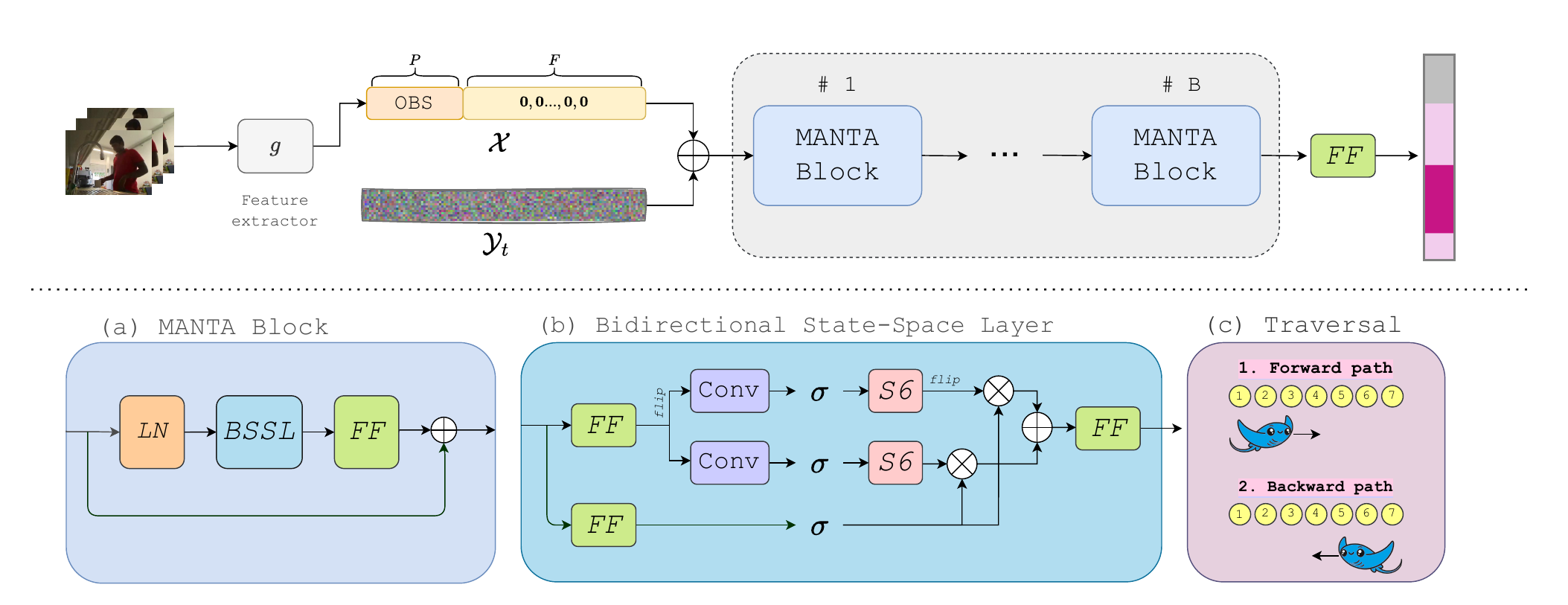}
    \vspace{-0.5cm}
    \caption{(\textit{Top}) Overview of the proposed MANTA model. Given a noise vector $\mathcal{Y}_t$ and a conditioning vector $\mathcal{X}$, constructed by extending the features of the $P$ observed frames with zero padding in place of the $F$ future frames, we concatenate and forward them through our proposed MANTA model. As output, MANTA predicts action classes for both observed and future frames. \textit{(Bottom)} Illustration of the structure of the (a) MANTA block, (b) Bidirectional State-Space Layer, as well as (c) traversal directions of the BSSL for the temporal input sequence.}
    \label{fig:main}
    \vspace{-0.3cm}
\end{figure*}

\subsection{Model}
\label{sec:model}
The overview of our proposed MANTA model is shown in Fig.~\ref{fig:main}.
 The key component of our model is the Bidirectional Selective State-Space Layer (BSSL), which enables effective and efficient long-range temporal modelling and supports selective sequence processing through data-dependent gating. These properties are essential for long-term dense anticipation, where framewise prediction of past and future actions requires reasoning over long sequences with a structured pattern: observed frames followed by masked future frames. Next, we describe our proposed BSS layer in Sec.~\ref{sec:bssl}, after detailing deep state-space models in Sec.~\ref{sec:state_space_models}, which serve as its foundation. We then discuss the overall structure of our proposed MANTA model in Sec.~\ref{sec:manta}.

\subsubsection{Deep Structured State-Space Models}
\label{sec:state_space_models}
Deep structured state-space models (SSMs) are a recently developed class of sequence models, inspired by continuous linear time-invariant state-space models, that define a mapping from an input sequence \( x(t) \in \mathbb{R} \) to an output sequence \( y(t) \in \mathbb{R}\) through a latent state \( h(t) \in \mathbb{R}^N \) using a first-order ordinary differential equation:
\begin{align}
    \label{eq:cont_ssm_1}
    h'(t) &= \textbf{A}h(t-1) + \textbf{B}x(t), \\
    \label{eq:cont_ssm_2}
    y(t) &= \textbf{C}h(t),
\end{align}
where $\textbf{A} \in \mathbb{R}^{N \times 1}$, $\textbf{B} \in \mathbb{R}^{N \times 1}$ and $\textbf{C} \in \mathbb{R}^{N \times 1}$ are state, input and output matrices, respectively.
Deep SSMs are defined by transforming the above formulation into a discrete-time learnable system. To this end, continuous-time equations~\eqref{eq:cont_ssm_1} and~\eqref{eq:cont_ssm_2} are discretized with a time-step parameter \( \Delta \) in the following way:
\begin{align}
    \label{eq:tr_1}
    h_t &= \bar{\textbf{A}} h_{t-1} + \bar{\textbf{B}} x_t, \\
    \label{eq:tr_2}
    y_t &= \textbf{C} h_t, \\
    \label{eq:discr_1}
    \bar{\textbf{A}} &= \exp(\Delta \textbf{A}), \\
    \label{eq:discr_2}
    \bar{\textbf{B}} &= \exp(\Delta \textbf{A})^{-1} (\exp(\Delta \textbf{A}) - I) \Delta \textbf{B}.
\end{align}
These discretized equations form the foundation for a deep SSM called \textit{S4}~\cite{gu2022efficientlyS4}, which defines a sequence-to-sequence transformation parametrized by (\(\textbf{A}\), \(\textbf{B}\), \(\textbf{C}\), and \(\Delta\)) following Eq.~\eqref{eq:tr_1} and~\eqref{eq:tr_2}. To compute the output of this model, its parameters are transformed into (\(\bar{\textbf{A}}\), \(\bar{\textbf{B}}\)) using Eq.~\eqref{eq:discr_1} and~\eqref{eq:discr_2}, after which output \(y\) is acquired either via linear recurrence or global convolution. 

While \textit{S4} models maintain a global receptive field, they do not implement selective value processing, since the parameters (\(\textbf{A}\), \(\textbf{B}\), \(\textbf{C}\), and \(\Delta\)) remain fixed over time. Recently, Gu~\etal~\cite{gu2023mamba} proposed an improved \textit{S6} layer by introducing the selection mechanism into the \textit{S4} model. To this end, they proposed to make parameters input-dependent by computing them as functions of \(x \in \mathbb{R}^{B \times T \times D}\), \ie, \(\textbf{B}(x) \in \mathbb{R}^{B \times T \times N}\), \(\textbf{C}(x) \in \mathbb{R}^{B \times T \times N}\), and \(\Delta(x) \in \mathbb{R}^{B \times T \times D}\). This increases the model’s adaptability, enabling data-driven information selection from both hidden states $h_t$ and input sequence entries $x_t$. 

This selection is controlled by the parameters of \textit{S6}, which 
can be interpreted to have specific functions within the model, as described in~\cite{gu2023mamba, han2024demystify}. Parameter \( \Delta \) can be viewed as an \textit{input gate}, controlling how much information from the input \( x_t \) gets propagated into the hidden state \( h_t \). The parameter \( \bar{\textbf{B}} \) similarly influences the degree of contribution of \( x_t \) to \( h_t \). Conversely, \( \bar{\textbf{A}} \) functions as a \textit{forget gate}, determining the extent of pruning of the previous hidden state \( h_{t-1} \). Finally, \( \textbf{C} \) controls how much information from \( h_t \) propagates into the output \( y_t \).

In this way, the data-dependent gating of the \textit{S6} adopts a similar role as the gated convolution of the GTAN, which adjusts the output of temporal convolutional layers using gating. While data-dependent processing also leads to increased capacity of \textit{S6}, its output can be efficiently computed using a parallel-scan algorithm.

\subsubsection{Bidirectional State-Space Layer}
\label{sec:bssl}
For BSSL, we make use of the above-described \textit{S6} layer as our temporal processing unit, due to its attractive properties of maintaining a locally-aware~\cite{han2024demystify} global receptive field, in contrast to GTAN, whose receptive field is limited by the dilation factor of its convolution layers. Apart from that, the selective modelling of the input sequence based on the data-dependent gating of the underlying SSM model is crucial for enabling MANTA to perform anticipation effectively. 

We base our layer on the Mamba block, proposed in ~\cite{gu2023mamba}, which incorporates the \textit{S6} layer into the  residual block for efficient and effective sequence processing. The general overview of BSSL's structure is depicted in Fig.~\ref{fig:main}(b). 
Following~\cite{gu2023mamba}, our proposed layer consists of two branches: the main branch (top) and the residual branch (bottom).
The residual branch implements a multiplicative residual connection within the layer. Specifically, it uses a linear projection layer to map the input feature \(F_t^{b}\) to a space with $2$-times expanded channels, followed by the activation function $\sigma$ (SiLU~\cite{Hendrycks2016GaussianEL, Ramachandran2017swish}), resulting in the vector \(R_t^{b}\):
\begin{align}
    R_t^{b} = \sigma(FF(F_t^{b})).
    \vspace{-0.2cm}
\end{align}
The main branch consists of two sub-branches: forward and backward. While the original Mamba block processes the input sequence solely in a forward direction, several recent works~\cite{li2025videomamba, zhu2024visionMamba,liu2024vmamba} have demonstrated the sub-optimality of this approach for visual sequences due to a lack of temporal awareness. To address this issue, following~\cite{li2025videomamba}, we extend the main branch with an additional backward pass to process the input sequence in reverse order (Fig.~\ref{fig:main}(c)). We analyzed the benefit of this extension in Sec.~\ref{sec:abl}. 

As shown in Fig.~\ref{fig:main}(b), the forward and backward paths are applied to the linearly projected input sequence separately.
Both paths follow the same flow: after a depth-wise convolution, its output is activated using $\sigma$ and is then processed by the \textit{S6} layer. After that, a residual connection is applied to the outputs of both paths independently. Importantly, before and after processing the sequence with the backward path, it is flipped in time. Formally:
\begin{align}
    W_t^{b} &= \text{S6}(\sigma(H^{K} * FF(F_t^{b}))), \\
    B_t^{b} &= \text{S6}(\sigma(H^{K} * FF(\overleftarrow{F_t^{b}}))), \\
    W_t^{b} &= W_t^{b} \odot R_t^{b}, \quad B_t^{b} = \overleftarrow{B_t^{b}} \odot R_t^{b},
\end{align}
Here $H^{K}$ is a 1D convolutional filter with size $K=3$, flipping across temporal dimension is denoted as $\leftarrow$, while $*$ and $\odot$ are convolution and element-wise multiplication, respectively. The output feature of the BSSL block is computed by accumulating the outputs of backward and forward paths as follows:
\begin{align}
    O_t^{b} = FF(W_t^{b} + B_t^{b}).
\end{align}

\subsubsection{MANTA }
\label{sec:manta}
Depicted in Fig.~\ref{fig:main} (top), our proposed network consists of $B$ MANTA blocks that process the input to predict future actions. Given pre-extracted observed frame features, we first pad them with zeros in place of future frames as described in Eq.~\eqref{eq:padding}, creating a conditioning vector \(\mathcal{X} \in \mathbb{R}^{(P+F) \times n_d}\). $\mathcal{X}$ is then concatenated with latent variables \(\mathcal{Y}_t\) along the channel dimension to form \(F_t \in \mathbb{R}^{(P + F) \times (n_c + n_d)}\), which is sequentially processed by MANTA blocks to generate the output feature \(\tilde{F}_t\), based on which the final prediction \(\tilde{\mathcal{Y}}_{0, t} \in \mathbb{R}^{(P+F) \times n_c}\) is made using an MLP layer. Here, $n_c$ and $n_d$ denote number of classes and number of feature channels, respectively.

The structure of individual MANTA blocks follows the design shown in Fig.~\ref{fig:main}(a). For block \(b\), its input feature \(F_t^{b}\) is first normalized using Layer Normalization (LN)~\cite{ba2016layer} and then passed through a BSSL, which performs temporal modelling of the input. Finally, a channel-mixing feed-forward layer is applied, followed by a residual connection. In summary, we can express the block’s operation in the following way:
\begin{align} F_t^{b+1} = FF(BSSL(LN(F_{t}^{b}))) + F_t^{b}. \end{align}

\subsubsection{Training and Inference}
For training our proposed MANTA model, we follow the established diffusion training approach~\cite{ho2020denoisingDiff}. Given a step $t$ sampled uniformly at random, we obtain $\mathcal{Y}_t$ using the forward diffusion process. We then pass $\mathcal{Y}_t$ alongside the conditioning vector $\mathcal{X}$ to our proposed MANTA model to obtain reconstruction $\tilde{\mathcal{Y}}_{0, t}$ of the ground-truth action sequence $\mathcal{Y}_0$. Finally, we compute the $\mathcal{L}_2$-loss between the predicted and the one-hot-encoded ground-truth action sequences:
\begin{align}
    \mathcal{L}_{rec} =  \mathbb{E}_{\mathcal{Y}_0, t, \epsilon_t} || \mathcal{Y}_0 - \tilde{\mathcal{Y}}_{0, t}||^2.
\end{align}
During inference, we generate multiple predictions ($S$) for a single set of observed frames $\mathcal{X}$. To do so, we sample multiple start noise vectors $\{\mathcal{Y}_{T,s}\}_{s=1}^{S} \sim \mathcal{N}(\mathbf{0}, \mathbf{I})$, which we subsequently denoise using our proposed MANTA model to obtain $\{\tilde{\mathcal{Y}}_{0, 1, s}\}_{s=1}^{S}$. To shorten the inference time, following~\cite{zatsarynna2024gtd}, we employ the DDIM~\cite{song2021denoising} sampling method and take $D < T$ out of $T$ diffusion steps.

\definecolor{lightblue}{rgb}{0.88, 0.93, 1.0}

\section{Experiments}

\subsection{Datasets}
To evaluate our proposed model, we use three datasets, commonly utilized for stochastic long-term anticipation tasks: Breakfast~\cite{Kuehne12}, Assembly101~\cite{sener2022assembly101} and 50Salads~\cite{Stein2013CombiningEA}.

\textbf{Breakfast}~\cite{Kuehne12} is a dataset of cooking videos in which human actors prepare 10 different breakfast meals in various kitchen environments. It comprises 1,712 videos that are densely annotated with temporal segments across 48 action classes. The dataset includes videos of varying durations, with the longest sequence lasting 10.8 minutes, which requires an anticipation horizon of up to 5.4 minutes. Following previous work, we evaluate our model using 4 predefined cross-validation splits and report the mean performance across these splits.

\textbf{Assembly101}~\cite{sener2022assembly101} is a large-scale dataset of toy assembly and disassembly videos captured from multiple exocentric and egocentric viewpoints, comprising a total of 4,321 videos. The video sequences are densely annotated with temporal action segments from 202 coarse action classes. The longest video duration for Assembly101 is over 25 minutes, requiring anticipation up to 12.5 minutes into the future. Following prior work~\cite{zatsarynna2024gtd}, we train our network on the training split and report evaluation results on the validation set, as the test set is not publicly available.

\textbf{50Salads}~\cite{Stein2013CombiningEA} is a dataset of salad preparation videos, consisting of 50 video sequences annotated with dense temporal segments from 17 action classes. The mean video duration is 6.4 minutes, with the longest video sequence lasting 10.1 minutes, which requires an anticipation horizon of up to 5.1 minutes. For evaluation, we report the performance of our model averaged over the 5 standard predefined cross-validation splits.

Following previous methods, we make use of the same pre-extracted frame features. Specifically, we use I3D features from~\cite{farha2020gcpr, gong2022future} for the Breakfast and 50Salads datasets, while for Assembly101 we rely on TSM features~\cite{Lin2020TSMTS} provided by~\cite{sener2022assembly101}. Additionally, to ensure a fair comparison with~\cite{zatsarynna2024gtd}, we closely follow their implementation. We provide further details in the supp. mat.

\subsection{Protocol and Evaluation}
In this work, we adopt the stochastic long-term dense anticipation protocol proposed by Farha \textit{et al.}~\cite{farha2019uaaa}. In this setting, the observation length and anticipation horizon are defined relative to the video duration. Specifically, for a video \( v \) with \( n_v \) total frames, the model observes \( P = \alpha n_v \) frames and then anticipates action labels for the subsequent \( F = \beta n_v \) frames, where \( \alpha \) and \( \beta \) represent the observation and anticipation percentages.

For evaluation, we also closely follow prior work~\cite{zatsarynna2024gtd, farha2019uaaa}. Specifically, for each observed video snippet, we sample $S=25$ predictions from our model. As our evaluation metrics, we report Mean and Top-1 MoC (Mean over Classes) accuracy: Mean MoC measures the average MoC across the $S$ generated samples, while Top-1 MoC reflects the MoC of the best-matching sample. Consistent with previous works, we report the results for $\alpha \in \{0.2, 0.3\}$ and $\beta \in \{0.1, 0.2, 0.3, 0.5\}$.

\begin{table}[]
\centering
\arrayrulecolor{gray}
\resizebox{\linewidth}{!}{%
\begin{tabular}{ll rrrr | rrrr }
\toprule

\multirow{2}{*}{MoC} & \multirow{2}{*}{Method} & \multicolumn{4}{c}{$\beta \ (\alpha=0.2)$} & \multicolumn{4}{c}{$\beta \ (\alpha=0.3) $} \\
\cline{3-10}
& & \textit{0.1} & \textit{0.2} & \textit{0.3} & \textit{0.5} & \textit{0.1} & \textit{0.2} & \textit{0.3} & \textit{0.5} \\

\midrule
\midrule

\multicolumn{10}{c}{Breakfast} \\
\midrule

\multirow{5}{*}{\makecell{Mean}} 

& Tri-gram~\cite{farha2019uaaa}
& 15.4 & 13.7 & 12.9 & 11.9 
& 19.3 & 16.6 & 15.8 & 13.9 \\

& UAAA~\cite{farha2019uaaa}  
& 15.7 & 14.0 & 13.3 & 13.0  
& 19.1 & 17.2 & 17.4 & 15.0   \\

& DiffAnt~\cite{zhong2023diffant}
& 24.7 & 22.9 & 22.1 & 22.3
& 30.9 & 30.2 & 28.9 & 27.5 \\

& GTDA ~\cite{zatsarynna2024gtd}
& 24.0 & 22.0 & 21.4 & 20.6 
& 29.1 & 26.8 & 25.3 & 24.2 \\

\rowcolor{lightblue}
\cellcolor{white} & \textbf{Ours}
& \textbf{27.7} & \textbf{25.3} & \textbf{24.6} & \textbf{23.8}
& \textbf{34.2} & \textbf{30.9} & \textbf{29.1} & \textbf{27.7} \\

\midrule

\multirow{5}{*}{\makecell{Top-1}} 

& Tri-gram~\cite{farha2019uaaa}
& - & - & - & -
& - & - & - & - \\

& UAAA~\cite{farha2019uaaa}                     
& 28.9 & 28.4 & 27.6 & 28.0 
& 32.4 & 31.6 & 32.8 & 30.8   \\

& DiffAnt~\cite{zhong2023diffant}
& 31.3 & 29.8 & 29.4 & 30.1
& 37.4 & 37.0 & 36.3 & 34.8 \\

& GTDA~\cite{zatsarynna2024gtd}                   
& 51.2 & 47.3 & 45.6 & 45.0
& 54.0 & 50.4 & 49.6 & 47.8 \\

\rowcolor{lightblue}
\cellcolor{white} & \textbf{Ours}
& \textbf{55.5} & \textbf{51.0} & \textbf{47.9} & \textbf{46.9} 
& \textbf{59.6} & \textbf{55.0} & \textbf{53.7} & \textbf{51.9} \\

\midrule
\multicolumn{10}{c}{Assembly101}\\
\midrule

\multirow{4}{*}{\makecell{Mean}}

& Tri-gram~\cite{zatsarynna2024gtd}
& 2.8 & 2.2 & 1.9 & 1.5 
& 3.5 & 2.7 & 2.3 & 1.8 \\

& UAAA~\cite{farha2019uaaa}
& 2.7 & 2.1 & 1.9 & 1.7 
& 2.4 & 2.1 & 1.9 & 1.7 \\

& GTDA~\cite{zatsarynna2024gtd} 
& 6.4 & 4.5 & 3.5 & 2.8 
& 5.9 & 4.2 & 3.5 & 2.9 \\

\rowcolor{lightblue}
\cellcolor{white} & \textbf{Ours}
& \textbf{6.7} & \textbf{5.3} & \textbf{4.2} & \textbf{3.5}
& \textbf{6.6} & \textbf{4.7} & \textbf{4.2} & \textbf{3.5} \\

\midrule

\multirow{4}{*}{\makecell{Top-1}}

& Tri-gram~\cite{zatsarynna2024gtd}
& 9.0 & 8.0 & 7.2 & 5.6 
& 9.5 & 8.2 & 7.8 & 5.9 \\

& UAAA~\cite{farha2019uaaa}  
& 6.9 & 5.9 & 5.6 & 5.1 
& 5.9 & 5.5 & 5.2 & 4.9 \\

& GTDA~\cite{zatsarynna2024gtd} 
& \textbf{18.0} & 12.8 & 9.9 & 7.7 
& \textbf{16.0} & 11.9 & 10.2 & 7.7 \\

\rowcolor{lightblue}
\cellcolor{white} & \textbf{Ours}
& 16.9 & \textbf{13.3} & \textbf{10.2} & \textbf{8.8}
& 15.6 & \textbf{12.0} & \textbf{11.1} & \textbf{8.4} \\

\midrule
\multicolumn{10}{c}{50Salads} \\
\midrule

\multirow{4}{*}{\makecell{Mean}}

& Tri-gram~\cite{farha2019uaaa}
& 21.4 & 16.4 & 13.3 & 9.4 
& 24.6 & 15.6 & 11.7 & 8.6  \\

& UAAA~\cite{farha2019uaaa}
& 23.6 & 19.5 & 18.0 & 12.8
& 28.0 & 18.0 & 14.8 & 12.1 \\

& GTDA~\cite{zatsarynna2024gtd} 
& 28.3 & 22.1 & 17.8 & 11.7 
& 29.9 & 18.5 & 14.2 & 10.6 \\

\rowcolor{lightblue}
\cellcolor{white} & \textbf{Ours}
& \textbf{28.6} & \textbf{22.8} & \textbf{19.5} & \textbf{13.6}
& \textbf{31.3} & \textbf{21.9} & \textbf{17.6} & \textbf{13.0} \\

\midrule

\multirow{4}{*}{\cellcolor{white} \makecell{Top-1}} 

& Tri-gram~\cite{farha2019uaaa}
& - & - & - & -
& - & - & - & -  \\

& UAAA~\cite{farha2019uaaa}  
& 53.5 & 43.0 & 40.5 & \textbf{33.7 }
& 56.4 & 42.8 & 35.8 & 30.2 \\

& GTDA~\cite{zatsarynna2024gtd}
& \textbf{69.6} & \textbf{55.8} & \textbf{45.2} & 28.1
& 66.2 & 44.9 & 39.2 & 31.0
\\

\rowcolor{lightblue}
\cellcolor{white} & \textbf{Ours}
& 68.3 & 51.5 & 41.7 & 31.3
& \textbf{71.7} & \textbf{53.3} & \textbf{43.8} & \textbf{31.1}
\\
\bottomrule
\end{tabular}
}
\vspace{-0.2cm}
\caption{Comparison of MANTA to state-of-the-art methods on Breakfast, Assembly101 and 50Salads.}
\label{tab:results_main}
\vspace{-0.3cm}
\end{table}

\subsection{Results}
\textbf{Comparison to State of the Art.} In this section, we present the results of our proposed model, comparing its performance with previous methods, as shown in Tab.~\ref{tab:results_main}. 
Our model achieves state-of-the-art results across all three datasets for both reported metrics. 
Specifically, on the Breakfast dataset, our method significantly outperforms previous approaches across all observation and anticipation horizons. On the Assembly101 dataset, MANTA surpasses all prior methods, with the exception of Top-1 MoC accuracy for \( \beta = 0.1 \), where it performs slightly worse than GTDA. Lastly, on the 50Salads dataset, our approach sets a new state of the art for Mean MoC accuracy and Top-1 MoC accuracy for \( \alpha = 0.3 \). For \( \alpha = 0.2 \), our model's Top-1 MoC is slightly lower compared to earlier methods. Generally, the average improvement in Top-1 MoC of MANTA compared to previous works across different anticipation ratios is consistently larger for $\alpha{=}0.3$ than for $\alpha{=}0.2$ for all datasets, indicating superior performance of our model for longer input sequences.

\begin{figure}[b!]
    \vspace{-0.4cm}
    \centering
        \includegraphics[width=0.49\columnwidth]{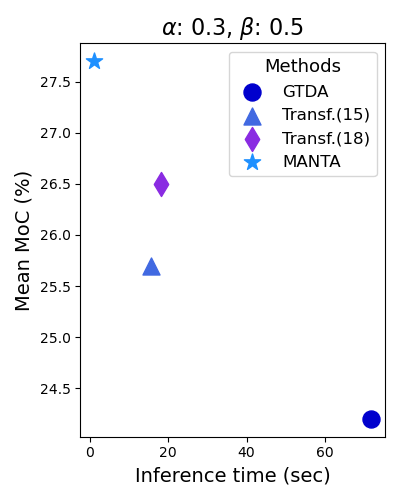}  
        \includegraphics[width=0.49\columnwidth]{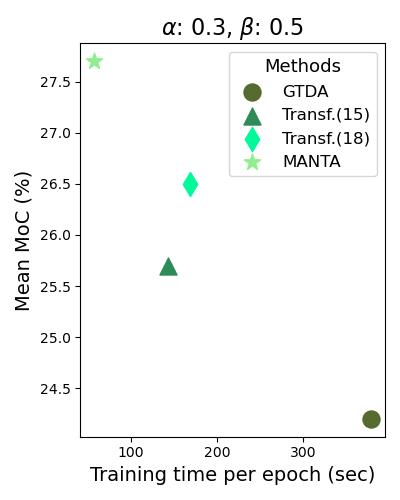}  
        \label{fig:image1}
    \vspace{-0.75cm}
    \caption{(\textit{Left}) Mean time for generating 25 samples measured for different models on Breakfast, with all models performing \(50\) inference diffusion steps; (\textit{Right}) Mean time required for training different models for one epoch on Breakfast. (\textit{Both}) The batch size is equal across models, and evaluation/training was conducted on the same GPU.}
    \vspace{-0.1cm}
    \label{fig:efficiency}
\end{figure}

Beyond its effectiveness, our proposed model demonstrates substantial efficiency improvements compared to~\cite{zatsarynna2024gtd}. Fig.~\ref{fig:efficiency} compares the inference and training time of MANTA and the previous best-performing method GTDA~\cite{zatsarynna2024gtd} on the Breakfast dataset. On the left, it shows the average inference time required to generate \( S=25 \) samples with observation and anticipation ratios set to \( \alpha = 0.3 \) and \( \beta = 0.5 \) for videos in the test splits. While GTDA requires an average of \( 71.8 \) seconds per video, our model needs only \( 1.1 \) seconds, achieving a \( 65.3 \times \) speedup. Besides faster inference, our model significantly reduces training time as well, as shown in Fig.~\ref{fig:efficiency} on the right. The average runtime for a single epoch of GTDA is 379 seconds, while MANTA only takes 57 seconds per epoch. Additionally, due to the reduced number of the required blocks, our proposed model has $2.8\times$ fewer parameters compared to GTDA, as shown in Tab.~\ref{tab:params} and is more memory-efficient.

\begin{figure*}[t!]
    \centering
    \includegraphics[width=0.9\linewidth]{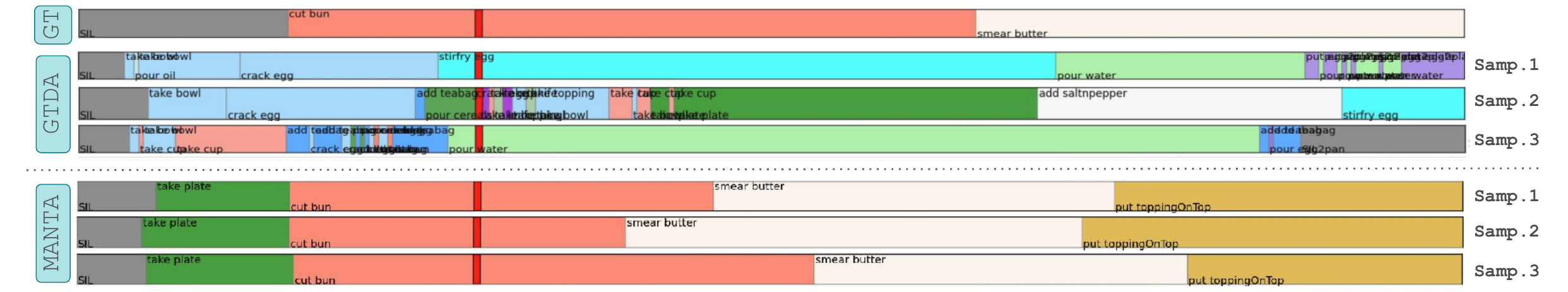}  
    \vspace{-0.4cm}
    \caption{Qualitative comparison of MANTA and GTDA on Breakfast. Best viewed zoomed in.}
    \label{fig:qual}
    \vspace{-0.3cm}
\end{figure*}

Finally, we show a qualitative comparison of our model to GTDA in Fig.~\ref{fig:qual}. In this example, GTDA anticipates future actions wrong, while our proposed MANTA model predicts plausible action sequences. Here, the plausibility is judged by the number of predicted segments and their corresponding classes. GTDA produces a high number of segments and includes actions from unrelated recipes (e.g., pour cereal and break egg).  This over-segmentation results from GTDA’s global but sparse receptive field of its later layers. In contrast, MANTA’s dense receptive field ensures consistency in the predicted segments and their classes. More qualitative results are in the supp.\ mat.

\subsection{Ablations}
In this section, we discuss a series of experiments that we performed to explore individual components of our proposed MANTA model. Further ablations experiments are provided in the supp. mat.

\label{sec:abl}

\begin{table}[b!]
\centering
\vspace{-0.1cm}
\resizebox{\columnwidth}{!}{%
\begin{tabular}{l c rrrr | rrrr }
\toprule
\multirow{2}{*}{MoC} & BSSL  & \multicolumn{4}{c}{$\beta \, (\alpha=0.2)$} & \multicolumn{4}{c}{$\beta \, (\alpha=0.3)$} \\
\cline{3-10}
& type & \textit{0.1} & \textit{0.2} & \textit{0.3} & \textit{0.5} & \textit{0.1} & \textit{0.2} & \textit{0.3} & \textit{0.5} \\
\midrule
\midrule
\multirow{2}{*}{\makecell{Mean}} 
& Ours w/o Sel.
& 26.4 & 24.4 & 23.5 & 23.2
& 31.7 & 28.9 & 27.3 & 26.5 \\
& Ours w/ Sel.
& \textbf{27.7} & \textbf{25.3} & \textbf{24.6} & \textbf{23.8}
& \textbf{34.2} & \textbf{30.9} & \textbf{29.1} & \textbf{27.7} \\
\midrule
\multirow{2}{*}{\makecell{Top-1}} 
& Ours w/o Sel.
& 47.3 & 44.5 & 41.9 & 41.8
& 51.6 & 47.5 & 46.3 & 44.6 \\
& Ours w/ Sel.
& \textbf{55.5} & \textbf{51.0} & \textbf{47.9} & \textbf{46.9} 
& \textbf{59.6} & \textbf{55.0} & \textbf{53.7} & \textbf{51.9} \\
\midrule
\end{tabular}
}
\vspace{-0.3cm}
\caption{Ablation of the selectivity of the BSSL layer on Breakfast.}
\vspace{-0.2cm}
\label{tab:select}
\end{table}

\textbf{Bidirectional State-Space Layer. }
The key component of MANTA is the BSS layer, discussed in Sec.~\ref{sec:bssl}. We conducted a series of experiments to investigate how different formulations of this layer affect the proposed model. Specifically, we examine how the following properties contribute: \textit{selectivity} and \textit{bi-directionality}.

Firstly, the \textit{selectivity} of the BSS layer is enabled by employing the \textit{S6} layer for temporal modelling, which applies input-specific SSM matrices for data-driven sequence processing. To assess the impact of selectivity, we replaced the \textit{S6} layer with \textit{S4}, for which the parameters remain fixed across timesteps. The results in Tab.~\ref{tab:select} show a significant performance drop without selectivity (Ours w/o Sel.). This is expected, as the input sequences to our model consist of two distinct segments: observed and masked future frames. Processing these segments homogeneously is inherently sub-optimal, since it is not possible to adjust the contribution of individual values regardless of their validity and/or importance. GTDA made similar observations, where selectivity was ensured by data-driven gating.

Secondly, we investigated the impact of \textit{bi-directionality} in the BSSL, which processes the input sequence from both ends using forward and backward branches (Fig.~\ref{fig:main}(b) and (c)). To assess the role of the backward branch, we tested two models where the backward path was removed. First, we removed it from the first \(10\) MANTA blocks, leaving only the last \(5\) bidirectional blocks, and then removed it entirely, resulting in \(0\) bidirectional blocks. The results in Tab.~\ref{tab:traverse} show a performance drop in both cases, with the performance of the fully uni-directional network particularly affected. This suggests that the backward flow of information is crucial for model performance, highlighting the importance of sequence processing in both directions.

\begin{table}[b!]
\vspace{-0.3cm}
\centering
\resizebox{\linewidth}{!}{%
\begin{tabular}{l cc rrrr | rrrr }
\toprule
\multirow{2}{*}{MoC} & Blocks & Blocks. & \multicolumn{4}{c}{$\beta (\alpha=0.2)$} & \multicolumn{4}{c}{$\beta (\alpha=0.3)$} \\
\cline{4-11}
& w/ Fwd. & w/ Bwd. & \textit{0.1} & \textit{0.2} & \textit{0.3} & \textit{0.5} & \textit{0.1} & \textit{0.2} & \textit{0.3} & \textit{0.5} \\
\midrule
\midrule

\multirow{3}{*}{\makecell{Mean}} 
& 15 & 15
& \textbf{27.7} & \textbf{25.3} & \textbf{24.6} & \textbf{23.8}
& \textbf{34.2} & \textbf{30.9} & \textbf{29.1} & \textbf{27.7} \\
& 15 & 5
& 26.8 & 24.8 & 24.3 & 23.1
& 32.5 & 29.8 & 28.0 & 26.4 \\
& 15 & 0
& 22.3 & 20.1 & 19.1 & 18.1
& 26.6 & 23.9 & 23.0 & 20.3 \\
\midrule

\multirow{3}{*}{\makecell{Top-1}} 
& 15 & 15
& \textbf{55.5} & \textbf{51.0} & \textbf{47.9} & \textbf{46.9}
& \textbf{59.6} & \textbf{55.0} & \textbf{53.7} & \textbf{51.9} \\
& 15 & 5
& 52.7 & 49.5 & 47.1 & 46.8
& 57.2 & 53.0 & 51.2 & 50.0 \\
& 15 & 0
& 35.8 & 33.7 & 32.4 & 31.2
& 39.6 & 37.2 & 36.2 & 32.7 \\
\midrule
\end{tabular}
}
\vspace{-0.3cm}
\caption{Ablation of the BSSL scanning directions on Breakfast.}
\vspace{-0.2cm}
\label{tab:traverse}
\end{table}

\begin{table}[b!]
\centering
 \resizebox{1.0\linewidth}{!}{%
\begin{tabular}{ll rrrr | rrrr }
\toprule

\multirow{2}{*}{MoC} & \multirow{2}{*}{Block} & \multicolumn{4}{c}{$\beta \ (\alpha=0.2) $} & \multicolumn{4}{c}{$\beta \ (\alpha=0.3) $} \\
\cline{3-10}
& & \textit{0.1} & \textit{0.2} & \textit{0.3} & \textit{0.5} & \textit{0.1} & \textit{0.2} & \textit{0.3} & \textit{0.5} \\

\midrule
\midrule

\multirow{4}{*}{\makecell{Mean}} 

&  Ours w/o FF
& 27.6 & 25.2 & 24.3 & 23.7
& 34.0 & 30.4 & 28.8 & \textbf{27.7}
\\

&  Ours w/ SE~\cite{Hu2017SqueezeandExcitationN}
& 26.5 & 24.2 & 23.7 & 23.1
& 33.3 & 30.0 & 28.2 & 27.3
\\

&  Ours w/ Dil.Conv
& 26.8 & 24.0 & 23.4 & 22.6
& 31.6 & 29.3 & 27.0 & 25.9
\\

& {Ours w/ \cite{li2025videomamba}}
& 26.7 & 24.2 & 23.4  & 22.6 
& 33.2 & 30.1 & 28.1 & 27.1 
\\

&  Ours
& \textbf{27.7} & \textbf{25.3} & \textbf{24.6} & \textbf{23.8}
& \textbf{34.2} & \textbf{30.9} & \textbf{29.1} & \textbf{27.7} \\

\midrule

\multirow{4}{*}{\makecell{Top-1}}

&  Ours w/o FF
& 52.2 & 48.9 & 46.7 & 46.6
& 54.9 & 50.3 & 47.8 & 42.4
\\

&  Ours w/ SE~\cite{Hu2017SqueezeandExcitationN}
& 51.6 & 48.9 & 47.0 & 46.1
& 57.5 & 53.1 & 52.4 & 50.2
\\

&  Ours w/ Dil. Conv
& 46.6 & 42.5 & 41.4 & 39.7
& 51.6 & 48.0 & 45.4 & 43.9
\\

&  {Ours w/ \cite{li2025videomamba}}
& \textbf{57.2} & \textbf{52.0} & \textbf{49.5} & \textbf{48.1} 
& \textbf{60.1} & \textbf{56.2} & 53.6  & \textbf{52.7} 
\\

& Ours
& {55.5} & {51.0} & {47.9} & {46.9} 
& {59.6} & {55.0} & \textbf{53.7} & {51.9} \\

\midrule
\end{tabular}
}
\vspace{-0.3cm}
\caption{Ablation of the MANTA block structure on Breakfast. }
\vspace{-0.2cm}
\label{tab:blocks}
\end{table}

\textbf{MANTA Block. } 
We further performed a series of experiments to explore how changes to the MANTA Block beyond the BSSL affect the performance of the final model.
Firstly, we investigated the effect of the internal structure of the block, as shown in Tab.~\ref{tab:blocks}. 
Following recent works~\cite{guo2025mambair, shi2024multiscale}, we experimented with modifying the channel mixing (FF) component of our block. Removing the mixing MLP altogether (Ours w/o FF) hinders the performance, as expected, due to a lack of channel-wise communication. However, introducing channel attention (Ours w/ SE), as proposed in~\cite{guo2025mambair}, fails to improve the results and leads to a performance drop.
Moreover, we experimented with introducing a dilated convolution layer with an exponentially growing dilation rate (Ours w/ Dil. Conv) before the LN layer, akin to~\cite{zatsarynna2024gtd}. This, however, led to the degradation of results. Finally, we evaluated the variant \cite{li2025videomamba}, which adds a residual block with normalization and MLP, as in a transformer block, after each block with BSSL. While it achieves better  Top-1 MoC, Mean MoC is lower.   

\begin{table}[t!]
\centering
\resizebox{\linewidth}{!}{%
\begin{tabular}{ll rrrr | rrrr }
\toprule
\multirow{2}{*}{MoC} & \multirow{2}{*}{Method} & \multicolumn{4}{c}{$\beta \ (\alpha=0.2)$} & \multicolumn{4}{c}{$\beta \ (\alpha=0.3)$} \\
\cline{3-10}
& & \textit{0.1} & \textit{0.2} & \textit{0.3} & \textit{0.5} & \textit{0.1} & \textit{0.2} & \textit{0.3} & \textit{0.5} \\
\midrule
\midrule
\multirow{4}{*}{\makecell{Mean}} 
& GTDA~\cite{zatsarynna2024gtd}
& 24.0 & 22.0 & 21.4 & 20.6
& 29.1 & 26.8 & 25.3 & 24.2 \\
& Transf. (15)~\cite{ltc2023bahrami}
& 25.6 & 23.3 & 22.5 & 21.9
& 30.5 & 28.1 & 26.4 & 25.7 \\
& Transf. (18)~\cite{ltc2023bahrami}
& 27.3 & 24.9 & 24.3 & 23.3
& 32.0 & 29.2 & 27.5 & 26.5 \\
\rowcolor{lightblue} \cellcolor{white} & \textbf{Ours}
& \textbf{27.7} & \textbf{25.3} & \textbf{24.6} & \textbf{23.8}
& \textbf{34.2} & \textbf{30.9} & \textbf{29.1} & \textbf{27.7} \\
\midrule
\multirow{4}{*}{\makecell{Top-1}} 
& GTDA~\cite{zatsarynna2024gtd}
& 51.2 & 47.3 & 45.6 & 45.0
& 54.0 & 50.4 & 49.6 & 47.8 \\
& Transf. (15)~\cite{ltc2023bahrami}
& 51.2 & 48.4 & 45.9 & 44.0
& 54.1 & 51.2 & 49.3 & 46.5 \\
& Transf. (18)~\cite{ltc2023bahrami}
& 52.7 & 49.1 & 47.0 & 45.2
& 57.2 & 52.3 & 50.5 & 48.2 \\
\rowcolor{lightblue} \cellcolor{white} & \textbf{Ours}
& \textbf{55.5} & \textbf{51.0} & \textbf{47.9} & \textbf{46.9}
& \textbf{59.6} & \textbf{55.0} & \textbf{53.7} & \textbf{51.9} \\
\midrule
\end{tabular}
}
\vspace{-0.3cm}
\caption{Ablation of the MANTA block type on Breakfast.}
\label{tab:results_transf}
\vspace{-0.1cm}
\end{table}

\begin{table}[t!]
\centering
\resizebox{1.0\linewidth}{!}{%
\begin{tabular}{l |cc | ccc | c }
\toprule

Method  & Params. (M) & Mem. (GB) & Stages & Blocks & Total Blocks & Inf. Time (sec) \\

\midrule
\midrule

GTDA~\cite{zatsarynna2024gtd} & 3.9 & 19.2 & 5 & 9 & 45 & 71.8
\\

Transf. (15)~\cite{ltc2023bahrami} & 1.2 & 11.3 & 1 & 15 & 15 & 15.6
\\

Transf. (18)~\cite{ltc2023bahrami} & 1.4 & 13.2 & 1 & 18 & 18 & 18.2
\\

\rowcolor{lightblue}
\textbf{Ours} & 1.4 & 10.2 & 1 & 15 & 15 & 1.1
\\

\midrule
\end{tabular}
}
\vspace{-0.3cm}
\caption{Specifications for different models on Breakfast. Inference time is reported for the longest observation and anticipation ratios ($\alpha=0.3$, $\beta=0.5$).}
\vspace{-0.2cm}
\label{tab:params}
\end{table}

\begin{figure}[t!]
    \centering
    \includegraphics[width=0.7\columnwidth]{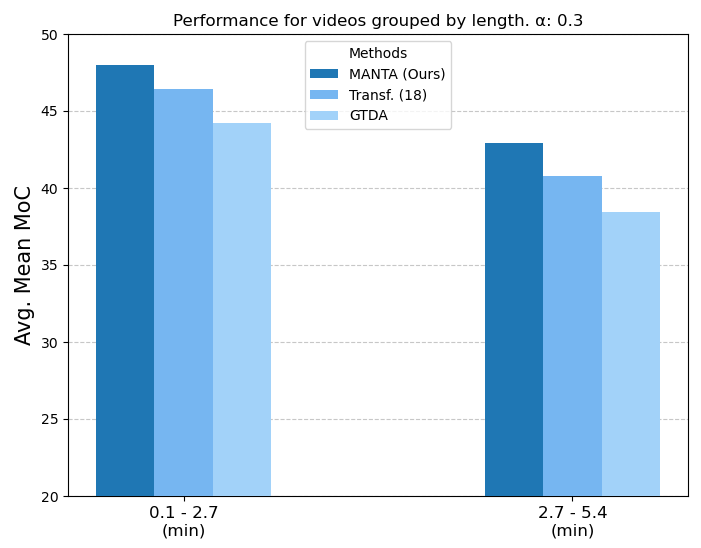}  
    \vspace{-0.4cm}
    \caption{Comparison of Average Mean MoC accuracy for different models on videos grouped by their duration on Breakfast for $\alpha=0.3$ and $\beta=0.2$. }
    \vspace{-0.3cm}
    \label{fig:length}
\end{figure}

\textbf{Attention. }
To better understand the benefits of the proposed model, particularly the state-space model-based MANTA block, we explored the alternative of utilizing the attention-based Transformer~(Transf.)~\cite{vaswani2017attention} block instead. Like SSM-based models, transformers have a global temporal receptive field, but unlike SSMs, each element in the sequence has direct access to all others, avoiding the need for information compression at the cost of quadratic complexity. To mitigate the high complexity, window-based transformers~\cite{ltc2023bahrami, liu2021Swin} constrain their computation to a limited set of values, making the computation sparse and therefore more efficient.

For our experiments, we made use of such a Transformer designed to balance efficiency and performance for long-term dense temporal action segmentation. Specifically, we use the block from~\cite{ltc2023bahrami}, which combines local windowed attention to ensure neighbouring consistency with global sparse `atrous' attention to model the sequence's long-term context. We compare our MANTA block to~\cite{ltc2023bahrami} in Tab.~\ref{tab:results_transf}, replacing all blocks in our network with theirs. We evaluate two Transformer models with different total block counts to ensure a fair comparison. As shown in Tab.~\ref{tab:params}, while Transf.(15) with $15$ blocks matches the block count in the MANTA architecture, Transf.(18) with $18$ blocks matches the parameter count of our proposed network. 
While both tested Transformer-based networks outperform the previously proposed GTDA in effectiveness and efficiency, among themselves, Transf.(18) is more effective than Transf.(15), but due to the larger number of layers, it suffers in efficiency. However, both Transformer models fall short compared to our proposed MANTA model. Specifically, our model is $16.5\times$ faster than Transf.(18) and achieves, on average, $1.1\%$ and $2.4\%$ better Mean and Top-1 MoC. 

\textbf{Long-Range Sequence Modelling. }
In Fig.~\ref{fig:length}, we further examine how our proposed MANTA model performs in comparison to the other tested models across different video lengths. Specifically, we calculate the Average Mean MoC for videos grouped by their duration, \ie, we first compute the Mean MoC for each video individually and then average the results within each length-based group.
While MANTA consistently outperforms both Transf.(18) and GTDA on all video lengths, the performance gap increases for longer videos. More specifically, MANTA shows a $3.8\%$ improvement in average Mean MoC over GTDA on shorter sequences, and a $5.5\%$ improvement on longer sequences.
As discussed in Sec.~\ref{sec:motivation}, GTDA encounters difficulties with longer sequences due to the lack of global context caused by the sparse receptive fields of its layers. Despite maintaining a global receptive field across layers, Transf.(18) also yields inferior results compared to our proposed model. We hypothesize that the issue stems from the lack of strong local bias in the Transformer’s temporal updates, which contrasts with the MANTA blocks. The lack of such bias would require the network to learn it from the data, potentially requiring specifically tailored optimization procedures and/or larger training sets.

\section{Conclusion}
\vspace{-0.2cm}
In this work, we proposed a novel MANTA model for stochastic long-term dense anticipation. While prior work aimed to model framewise past and future actions simultaneously, its performance was limited by the sparse receptive fields of its temporal modelling layers, resulting in redundant parameters and increased computational costs. To address these challenges, we introduced the MANTA generator network, which maintains a locally-aware global receptive field throughout the network, eliminating the need for multi-stage architectures and reducing both memory and computational requirements. Additionally, our model can adaptively process sequences with distinct observed and masked regions even for very long sequences. Experimental results on three datasets demonstrate that MANTA achieves state-of-the-art performance while achieving significant speed improvements.

\section*{Acknowledgments}
The work has been supported by the ERC Consolidator Grant FORHUE (101044724), the Deutsche Forschungsgemeinschaft
(DFG, German Research Foundation) GA 1927/4-2 (FOR 2535 Anticipating Human Behavior), and the project iBehave (receiving funding from the programme “Netzwerke 2021”, an initiative of the Ministry of Culture and Science of the State of NorthrhineWestphalia). The authors gratefully acknowledge EuroHPC Joint Undertaking for awarding us access to Leonardo at CINECA, Italy, through EuroHPC Regular Access Call- proposal No. EHPC-REG-2024R01-076. The authors also gratefully acknowledge the granted access to the Marvin cluster hosted by the University of Bonn. The sole responsibility for the content of this publication lies with the authors. We want to thank Mohamad Hakam Shams Eddin for his assistance with creating the teaser figure. The manta icons used in our paper were generated using Dall-E.
\clearpage
\setcounter{page}{1}
\maketitlesupplementary
Here, we present additional dataset and implementation details, as well as additional quantitative and qualitative results for our proposed MANTA model. More precisely, we discuss the implementation of our model in Sec.~\ref{sec:supp_impl} and provide additional details about the utilized datasets in Sec.~\ref{sec:datasets}. Next, in Sec.~\ref{sec:supp_abl} we present additional ablation studies for MANTA. Finally, in Sec.~\ref{sec:supp_qual} we present additional qualitative comparisons of MANTA to previous work.

\section{Implementation details}
\label{sec:supp_impl}
We implemented our model using Pytorch. As per Tab.~\ref{tab:num_blocks}, we use a total of $B=15$ MANTA blocks for our final model. Our proposed network is trained for 90 epochs using the Adam~\cite{kingma2014adam} optimizer with a learning rate of $0.0005$ for Breakfast and Assembly, and $0.001$ for 50Salads. Following~\cite{zatsarynna2024gtd}, we use $T=1000$ diffusion steps for training, $D=50$ DDIM steps for inference on Breakfast and Assembly101, and $D=10$ inference steps for 50Salads. 
\section{Datasets}
\label{sec:datasets}
In Tab.~\ref{tab:stats}, we show additional details for the datasets used in our work. Specifically, we provide average and maximum video durations, as well as the average and maximum number of individual segments per video. Since in our adapted anticipation protocol only up to $50\%$ of the video frames are utilized for anticipation, we additionally provide the statistics for the corresponding intervals of the videos in brackets in blue, including only frames falling into the anticipation intervals. As one can observe, the long temporal horizon of videos used for future anticipation and the numerous action segments that need to be predicted highlight the long-term nature of the addressed task.

\begin{table}[h!]
\centering
\resizebox{1.0\linewidth}{!}{%
\begin{tabular}{l |cc | cc }
\toprule

Dataset  & Avg. Num. Seg. & Max. Num. Seg & Avg. Dur. (min) & Max. Dur. (min) \\

\midrule
\midrule

Breakfast &
7 \textcolor{cvprblue}{(3)} & 25 \textcolor{cvprblue}{(15)} &
2.3 \textcolor{cvprblue}{(1.2)} & 10.8 \textcolor{cvprblue}{(5.4)}
\\

50Salads &
20 \textcolor{cvprblue}{(11)} & 26 \textcolor{cvprblue}{(18)} &
6.4 \textcolor{cvprblue}{(3.2)} & 10.1 \textcolor{cvprblue}{(5.1)}
\\

Assembly101 &
12 \textcolor{cvprblue}{(5)} & 73 \textcolor{cvprblue}{(40)} & 3.5 \textcolor{cvprblue}{(1.8)}
& 25.0 \textcolor{cvprblue}{(12.5)}
\\

\midrule
\end{tabular}
}
\vspace{-0.4cm}
\caption{\textit{(Left)} Number of segments and \textit{(Right)} duration for the whole video and in the \textcolor{cvprblue}{anticipation interval}.}
\label{tab:stats}
\vspace{-0.2cm}
\end{table}

\section{Ablation Study}
\label{sec:supp_abl}
\subsection{Bidirectional State-Space Layer (BSSL)}
In Tab.~2 and Tab.~3 of the main paper, we analyzed the \textit{selectivity} and \textit{bi-directionality} of the proposed BSS layer. Here, we examine how \textit{independent forward and backward} scanning contributes to the final performance of the MANTA model. More specifically, we tested if having \textit{independent} parameters for \textit{forward and backward} scanning branches is the best way to structure the BSS layer. To investigate this, we evaluated the effect of weight-sharing between the two branches. As shown in Tab.~\ref{tab:share}, while Top-1 MoC accuracy is similar across networks with shared and independent weights, Mean MoC accuracy is higher in the model with branch-specific weights. We, therefore, keep the weights separate for the two BSSL branches.

\begin{table}[h!]
\centering
\resizebox{1.0\linewidth}{!}{%
\begin{tabular}{l c rrrr | rrrr }
\toprule
\multirow{2}{*}{MoC} & \multirow{2}{*}{Shared} & \multicolumn{4}{c}{$\beta \ (\alpha=0.2)$} & \multicolumn{4}{c}{$\beta \ (\alpha=0.3)$} \\
\cline{3-10}
& & \textit{0.1} & \textit{0.2} & \textit{0.3} & \textit{0.5} & \textit{0.1} & \textit{0.2} & \textit{0.3} & \textit{0.5} \\
\midrule
\midrule
\multirow{2}{*}{\makecell{Mean}} 
& \textcolor{cvprblue}{\checkmark} & \textbf{28.5} & 23.4 & 23.8 & 22.9 & 33.2 & 27.7 & 28.2 & 27.1 \\
& \textcolor{lavender}{\ding{55}} & 27.7 & \textbf{25.3} & \textbf{24.6} & \textbf{23.8} & \textbf{34.2} & \textbf{30.9} & \textbf{29.1} & \textbf{27.7} \\
\midrule
\multirow{2}{*}{\makecell{Top-1}} 
& \textcolor{cvprblue}{\checkmark} & 53.7 & \textbf{51.0} & \textbf{48.8} & 46.7 & \textbf{60.7} & \textbf{55.9} & 53.3 & 50.5 \\
& \textcolor{lavender}{\ding{55}} & \textbf{55.5} & \textbf{51.0} & 47.9 & \textbf{46.9} & 59.6 & 55.0 & \textbf{53.7} & \textbf{51.9} \\
\midrule
\end{tabular}
}
\vspace{-0.2cm}
\caption{Ablation of weight sharing for forward and backward branches of the BSSL on Breakfast. }
\vspace{-0.2cm}
\label{tab:share}
\end{table}

\subsection{MANTA Block}
We experimented with varying the total number of blocks in the final model (Tab.~\ref{tab:num_blocks}). Empirically, we found that the model with \( B = 15 \) blocks showed the best results, with 
further increase or decrease in the number of blocks harming the model's performance.

\begin{table}[h!]
\centering

\resizebox{1.0\linewidth}{!}{%
\begin{tabular}{l c rrrr | rrrr }
\toprule

\multirow{2}{*}{MoC} & Num.  & \multicolumn{4}{c}{$\beta \ (\alpha=0.2) $} & \multicolumn{4}{c}{$\beta \ (\alpha=0.3) $} \\
\cline{3-10}
& blocks & \textit{0.1} & \textit{0.2} & \textit{0.3} & \textit{0.5} & \textit{0.1} & \textit{0.2} & \textit{0.3} & \textit{0.5} \\

\midrule
\midrule

\multirow{3}{*}{\makecell{Mean}}

& 10
& 26.6 & 24.3 & 23.5 & 23.1
& 32.7 & 29.4 & 28.1 & 26.8
\\

& 15
& \textbf{27.7} & \textbf{25.3} & \textbf{24.6} & \textbf{23.8}
& \textbf{34.2} & \textbf{30.9} & \textbf{29.1} & \textbf{27.7}
\\

& 20 
& 27.2 & 24.5 & 23.4 & 23.2
& 32.1 & 29.3 & 27.7 & 26.6
\\

\midrule

\multirow{3}{*}{\makecell{Top-1}}

& 10                     
& 54.2 & 49.2 & 46.7 & 46.6
& 57.2 & 52.5 & 52.4 & 49.7
\\

& 15
& \textbf{55.5} & \textbf{51.0} & 47.9 & 46.9
& \textbf{59.6} & \textbf{55.0} & \textbf{53.7} & \textbf{51.9}
\\

& 20                   
& 54.7 & 50.5 & \textbf{48.4} & \textbf{47.1}
& 57.8 & 53.4 & 52.3 & 50.5
\\

\midrule
\end{tabular}
}
\vspace{-0.2cm}
\caption{
Ablation of the number of MANTA blocks on Breakfast.
}
\vspace{-0.2cm}
\label{tab:num_blocks}
\end{table}


\subsection{Samples. }

We analyze the effect of the total sample count on MANTA's performance in Tab.~\ref{tab:num_samples}. The number of samples has a marginal effect on the Mean MoC accuracy whereas the Top-1 MoC increases with the number of samples. While this is expected, as Top-1 MoC only considers the sample that is closest to the ground-truth, the increase in Top-1 MoC with a higher number of samples demonstrates the diversity of the generated predictions. Otherwise, the Top-1 MoC would saturate after a small number of samples.

\begin{table}[h!]
\centering
\resizebox{1.0\linewidth}{!}{%
\begin{tabular}{ll rrrr | rrrr }
\toprule
\multirow{2}{*}{MoC} & Num.  & \multicolumn{4}{c}{$\beta \ (\alpha=0.2)$} & \multicolumn{4}{c}{$\beta \ (\alpha=0.3)$} \\
\cline{3-10}
& samples & \textit{0.1} & \textit{0.2} & \textit{0.3} & \textit{0.5} & \textit{0.1} & \textit{0.2} & \textit{0.3} & \textit{0.5} \\
\midrule
\midrule

\multirow{3}{*}{\makecell{Mean}} 
& 5  & \textbf{27.7} & \textbf{25.6} & \textbf{24.7} & \textbf{23.9} & 33.9 & 30.7 & 28.9 & 27.5 \\
& 15 & \textbf{27.7} & 25.4 & 24.6 & 23.8 & \textbf{34.2} & \textbf{30.9} & 29.0 & \textbf{27.8} \\
& 25 & 25.5 & 25.3 & 24.6 & 23.8 & \textbf{34.2} & \textbf{30.9} & \textbf{29.1} & 27.7 \\
\midrule

\multirow{3}{*}{\makecell{Top-1}} 
& 5  & 42.3 & 39.3 & 37.1 & 36.1 & 47.5 & 44.2 & 41.4 & 40.4 \\
& 15 & 51.4 & 47.3 & 44.7 & 44.0 & 55.8 & 52.0 & 50.5 & 48.6 \\
& 25 & \textbf{55.5} & \textbf{51.0} & \textbf{47.9} & \textbf{46.9} & \textbf{59.6} & \textbf{55.0} & \textbf{53.7} & \textbf{51.9} \\
\midrule
\end{tabular}
}
\vspace{-0.2cm}
\caption{Ablation of the number of samples on Breakfast.}
\vspace{-0.2cm}
\label{tab:num_samples}
\end{table}

\subsection{Robustness}
We report the standard deviation for the MANTA model, computed over 3 seeds on the Breakfast dataset, in Tab.~\ref{tab:std}. As expected, the std.\ is higher for Top-1 MoC, but the values are low, indicating the robustness of our proposed model.

\begin{table}[h!]
\centering
 \resizebox{0.85\linewidth}{!}{%
\begin{tabular}{l rrrr | rrrr }
\toprule

\multirow{2}{*}{MoC} & \multicolumn{4}{c}{$\beta \ (\alpha=0.2) $} & \multicolumn{4}{c}{$\beta \ (\alpha=0.3) $} \\
\cline{2-9}
& \textit{0.1} & \textit{0.2} & \textit{0.3} & \textit{0.5} & \textit{0.1} & \textit{0.2} & \textit{0.3} & \textit{0.5} \\

\midrule
\midrule

\multirow{1}{*}{\makecell{Mean}} 
& 0.07 & 0.07 & 0.06 & 0.05
& 0.04 & 0.18 & 0.09 & 0.11
\\

\midrule

\multirow{1}{*}{\makecell{Top-1}} 

& 0.5 & 0.6 & 0.3 & 0.2
& 0.3 & 0.6 & 0.8 & 0.5
\\
\bottomrule
\end{tabular}
}
\vspace{-0.3cm}
\caption{Standard deviation of MANTA on Breakfast dataset computed over 3 runs with different seeds.}
\vspace{-0.4cm}
\label{tab:std}
\end{table}

\section{Qualitative Results}
\label{sec:supp_qual}
\begin{figure*}[t!]
    \centering
    \includegraphics[width=0.75\textwidth]{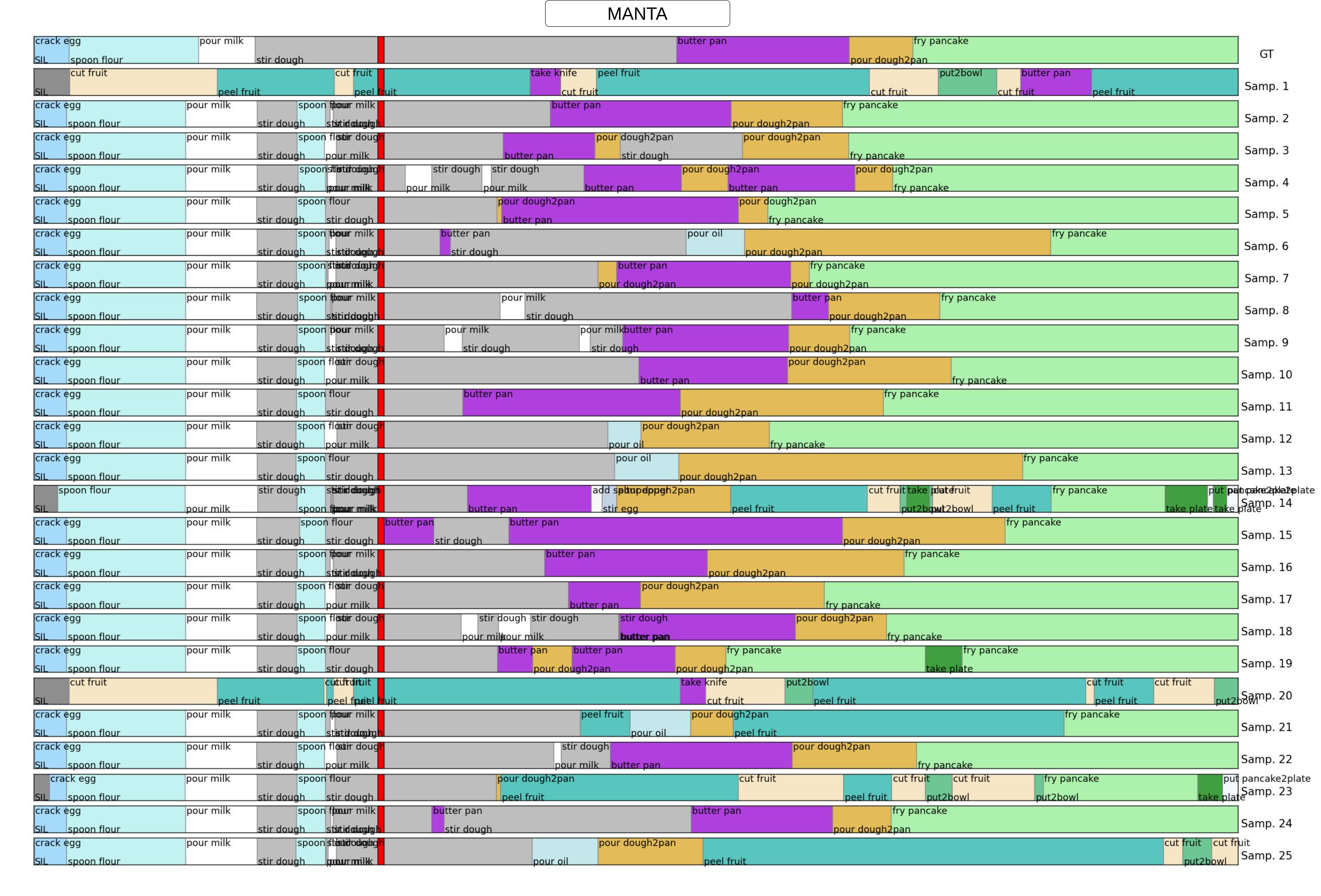}
    \includegraphics[width=0.77\textwidth]{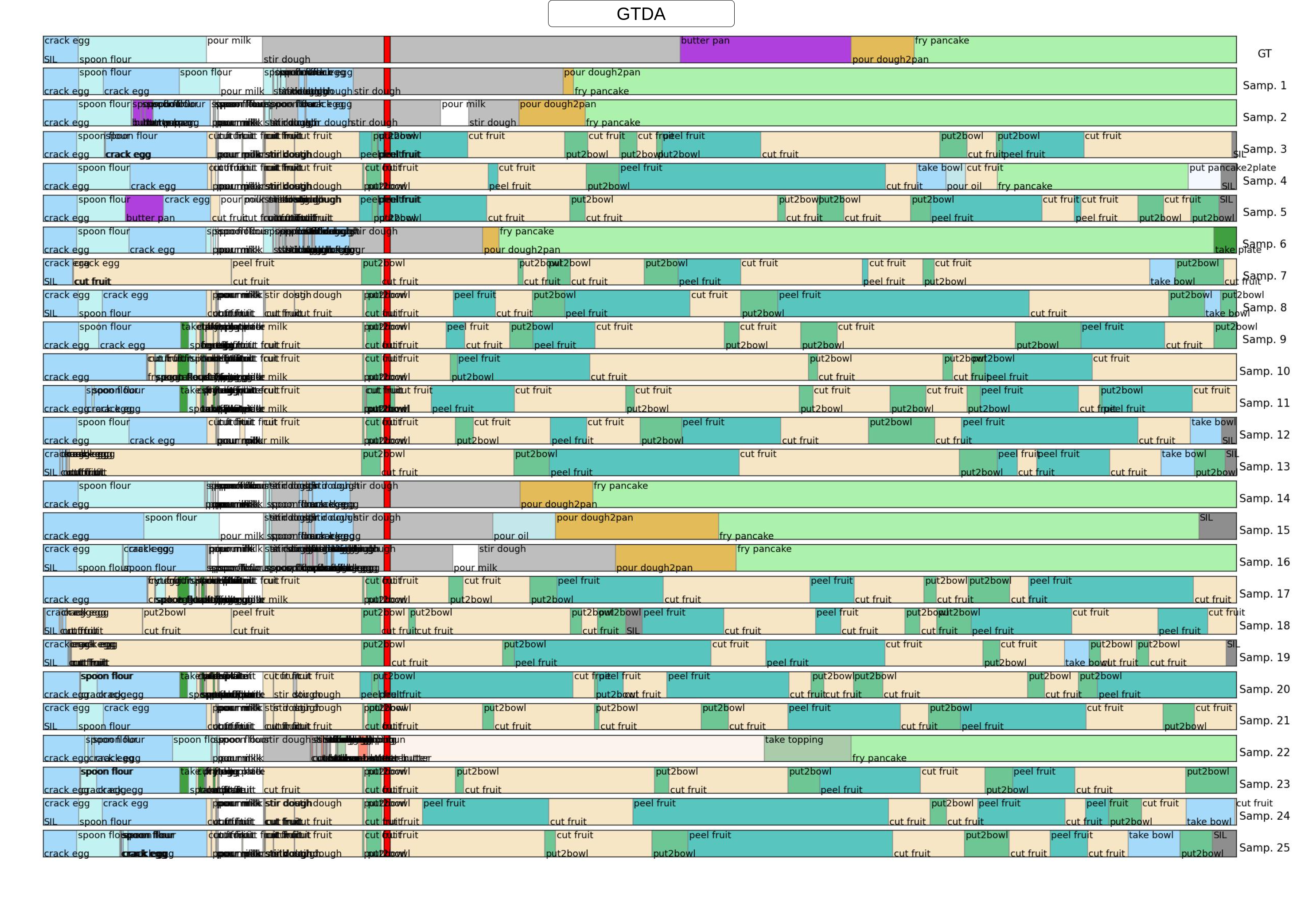}
    \caption{Qualitative comparison of MANTA \textit{(top)} to GTDA~\cite{zatsarynna2024gtd} \textit{(bottom)} on the Breakfast dataset. Best viewed zoomed in.}
    \label{fig:sup_qual_1}
\end{figure*}

\begin{figure*}[t!]
    \centering
    \includegraphics[width=0.75\textwidth]{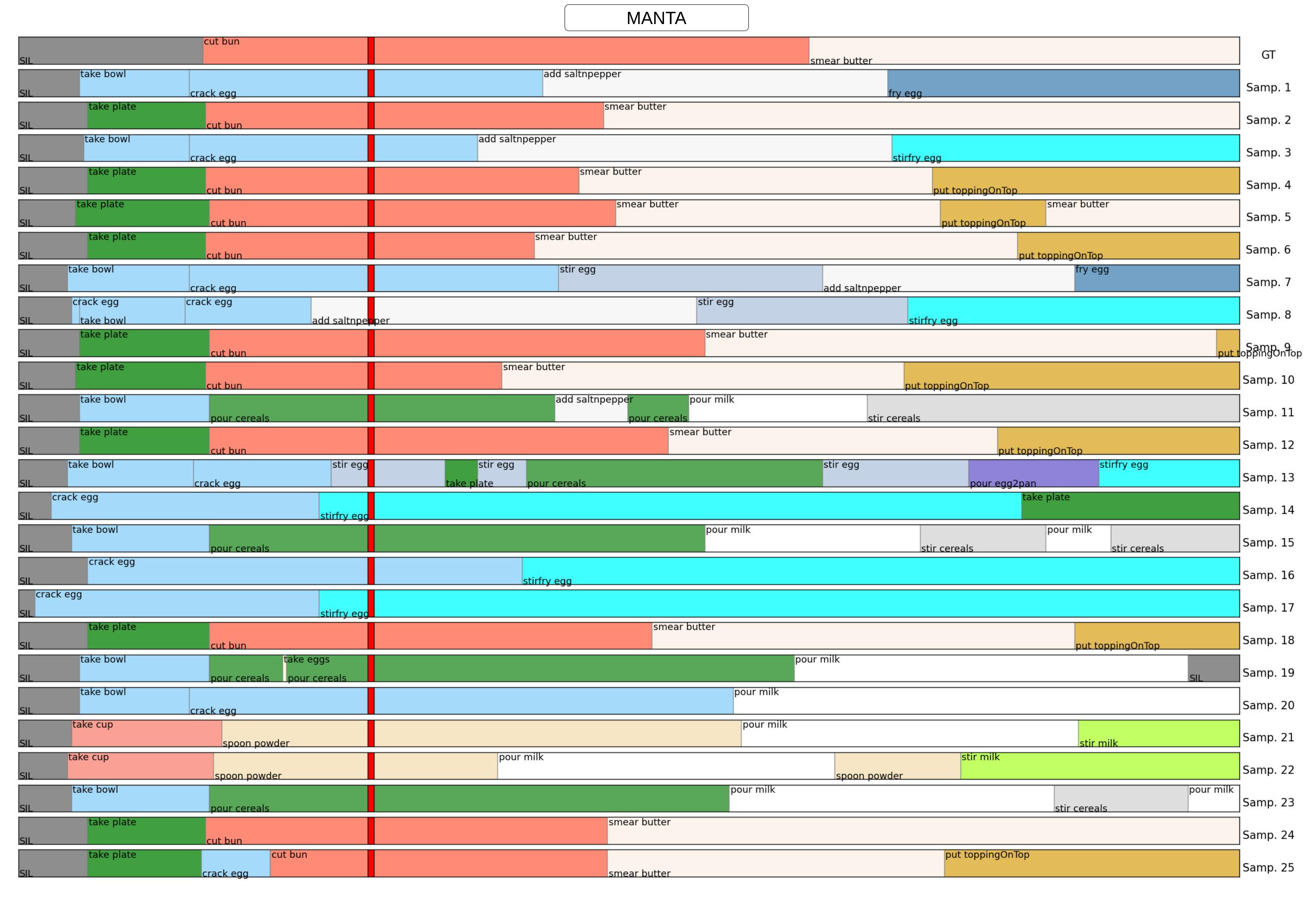}
    \includegraphics[width=0.76\textwidth]{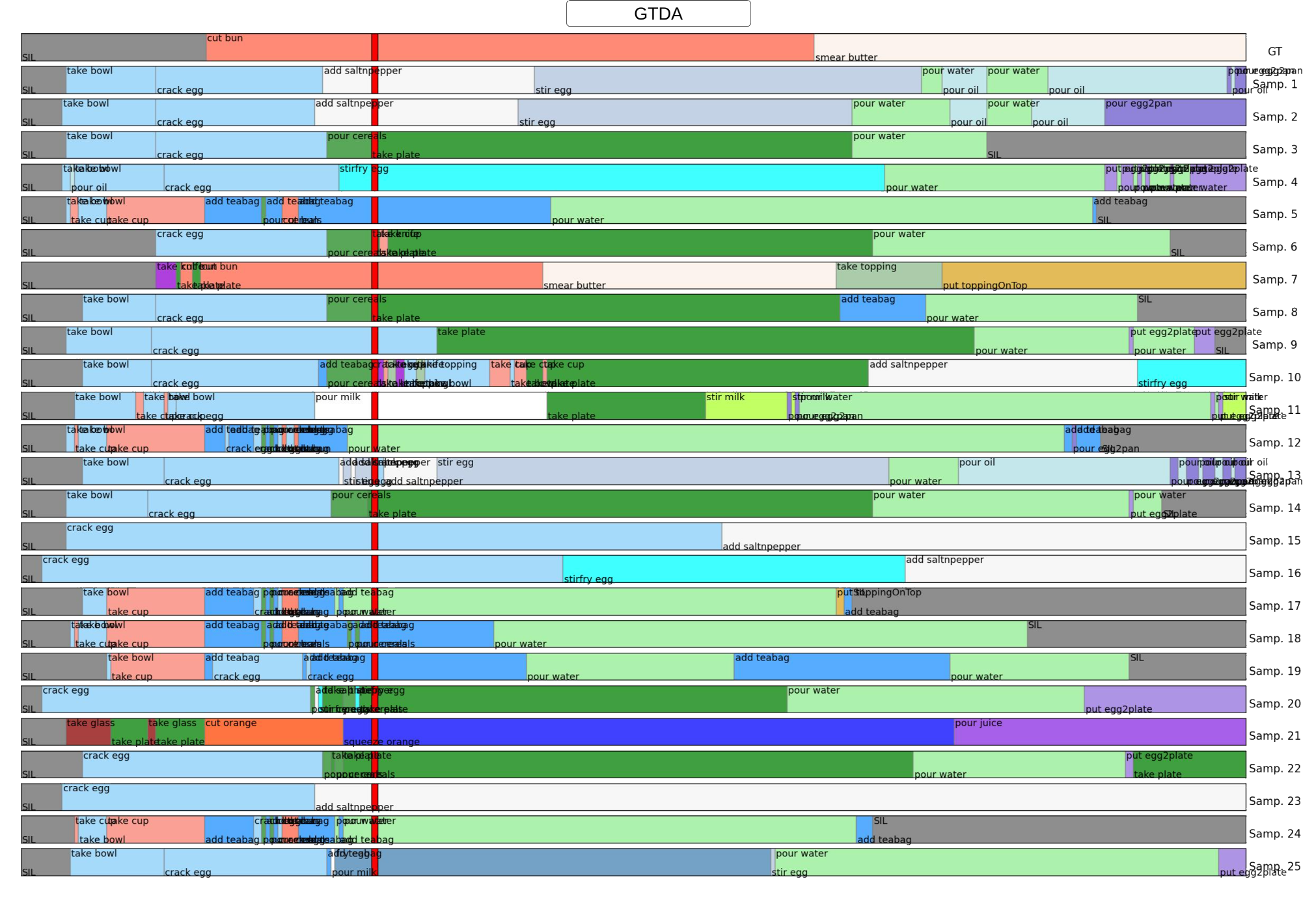}
    \caption{Qualitative comparison of MANTA \textit{(top)} to GTDA~\cite{zatsarynna2024gtd} \textit{(bottom)} on the Breakfast dataset. Best viewed zoomed in.}
    \label{fig:sup_qual_2}
\end{figure*}

\begin{figure*}[t!]
    \centering
    \includegraphics[width=0.75\textwidth]{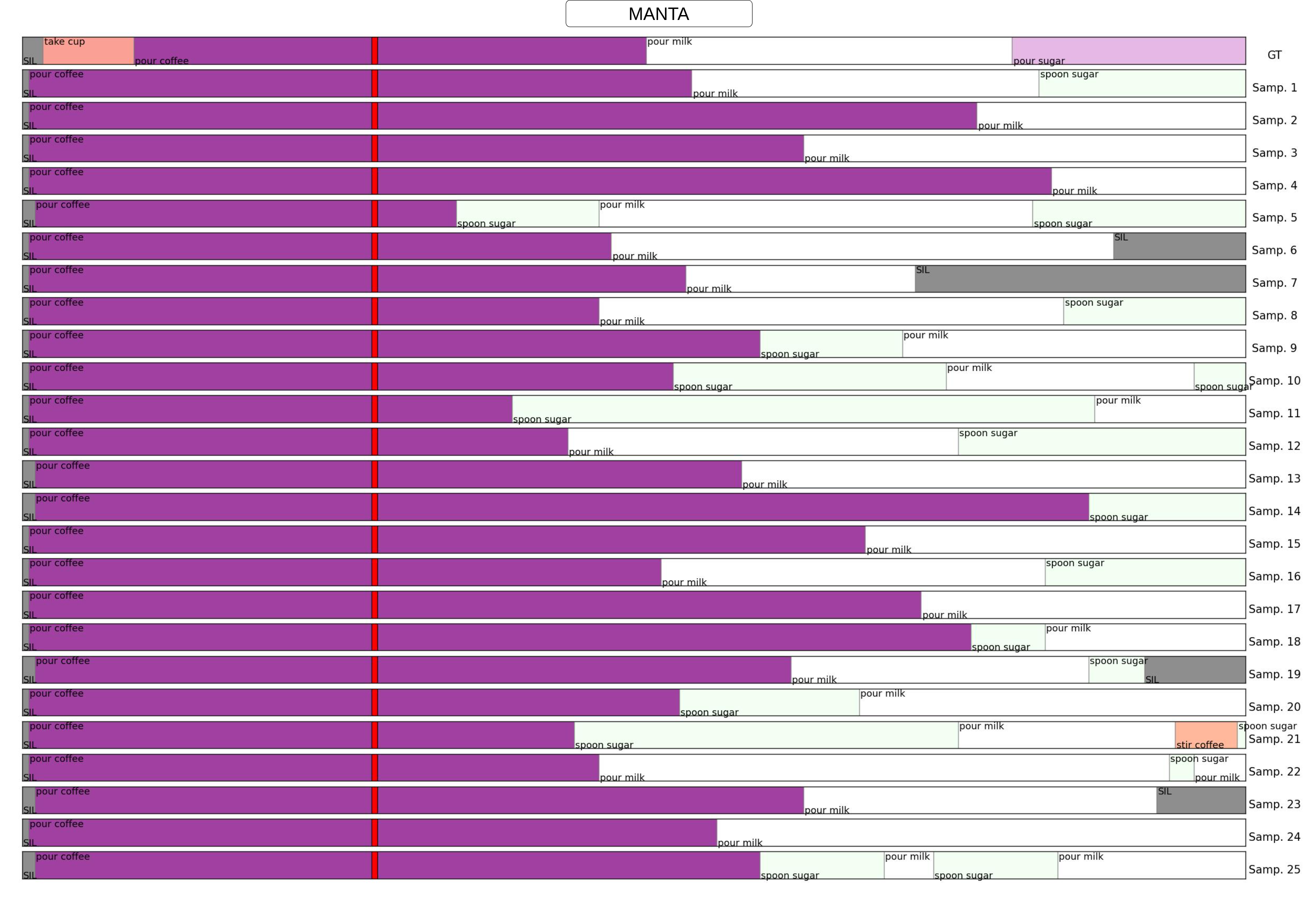}
    \includegraphics[width=0.753\textwidth]{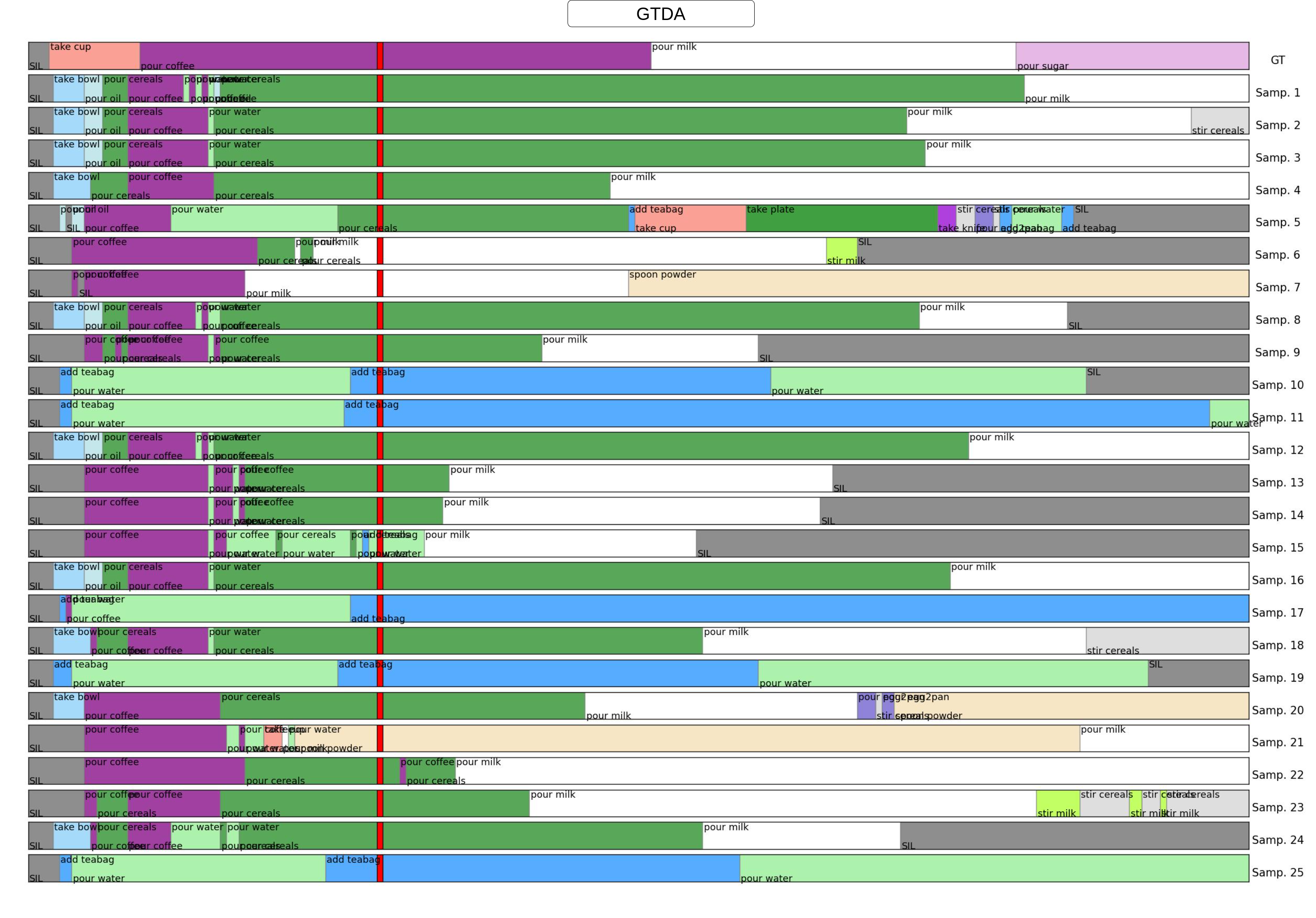}
    \caption{Qualitative comparison of MANTA \textit{(top)} to GTDA~\cite{zatsarynna2024gtd} \textit{(bottom)} on the Breakfast dataset. Best viewed zoomed in.}
    \label{fig:sup_qual_3}
\end{figure*}

\begin{figure*}[t!]
    \centering
    \includegraphics[width=0.95\textwidth]{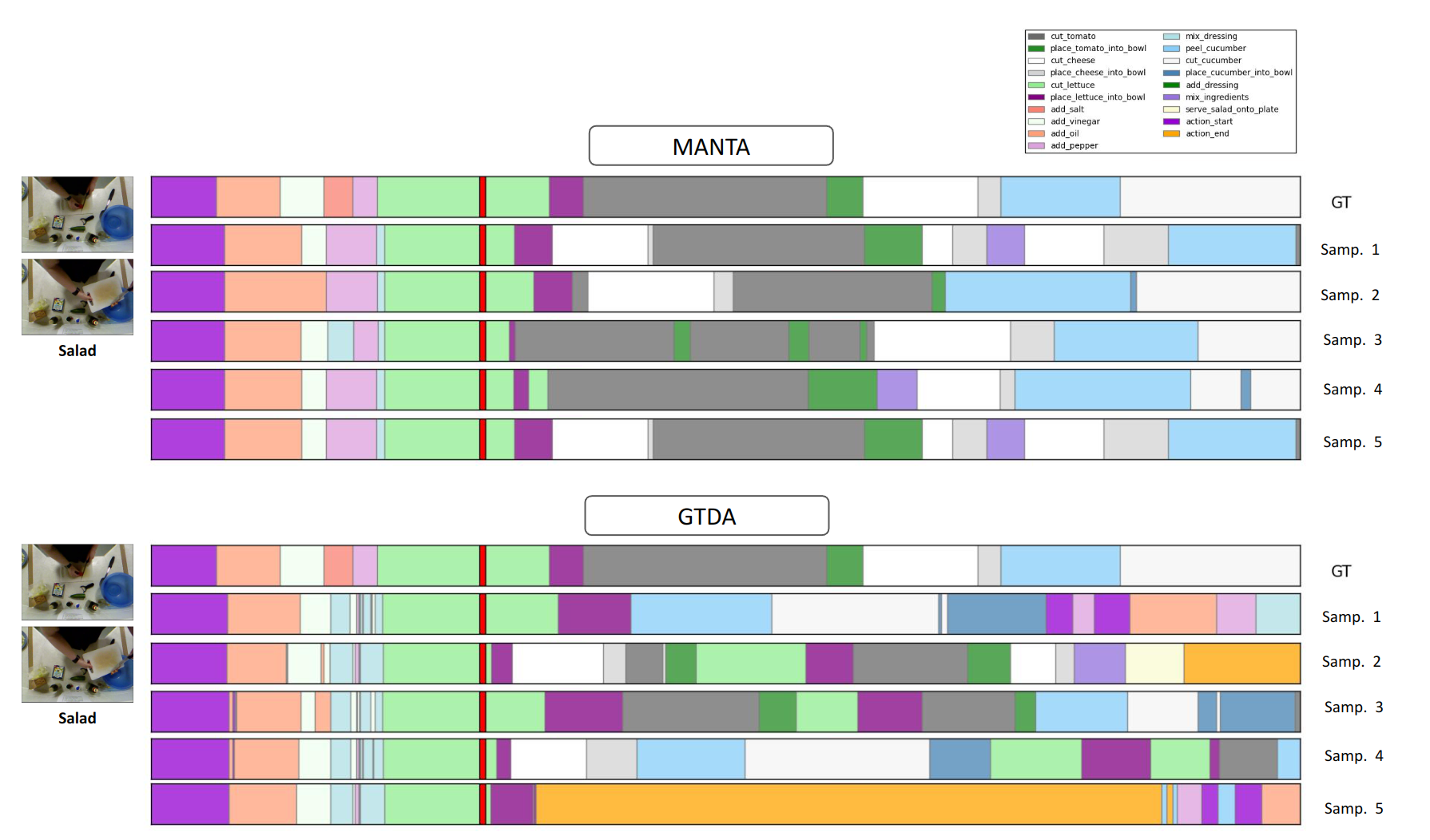}
    \caption{Qualitative comparison of MANTA \textit{(top)} to GTDA~\cite{zatsarynna2024gtd} \textit{(bottom)} on the 50Salads dataset. Best viewed zoomed in.}
    \label{fig:sup_qual_4}
\end{figure*}

\begin{figure*}[t!]
    \centering
    \includegraphics[width=0.95\textwidth]{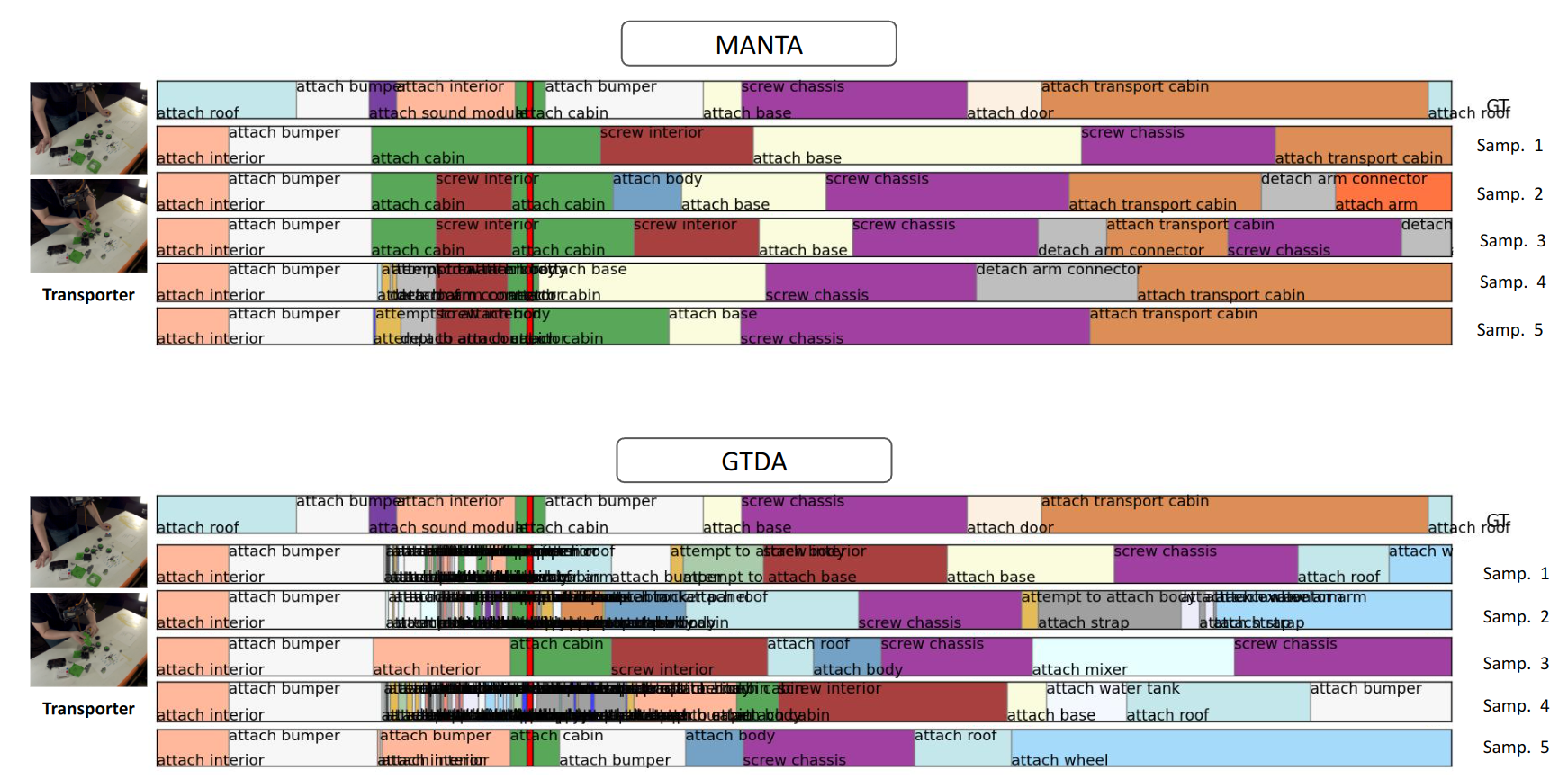}
    \caption{Qualitative comparison of MANTA \textit{(top)} to GTDA~\cite{zatsarynna2024gtd} \textit{(bottom)} on the Assembly101 dataset. Best viewed zoomed in.}
    \label{fig:sup_qual_5}
\end{figure*}

We provide qualitative comparisons of our proposed MANTA model to the previous best-performing GTDA~\cite{zatsarynna2024gtd} on the Breakfast dataset in Figs.~\ref{fig:sup_qual_1}-\ref{fig:sup_qual_3}, on the 50Salads dataset in Fig.~\ref{fig:sup_qual_4} and on Assembly101 dataset in Fig.~\ref{fig:sup_qual_5}.

{
	\clearpage
    \small
    \bibliographystyle{ieeenat_fullname}
    \bibliography{main}

\begin{thebibliography}{66}
\providecommand{\natexlab}[1]{#1}
\providecommand{\url}[1]{\texttt{#1}}
\expandafter\ifx\csname urlstyle\endcsname\relax
  \providecommand{\doi}[1]{doi: #1}\else
  \providecommand{\doi}{doi: \begingroup \urlstyle{rm}\Url}\fi

\bibitem[Abu~Farha et~al.(2018)Abu~Farha, Richard, and Gall]{Farha_2018_CVPR}
Y. Abu~Farha, A. Richard, and J. Gall.
\newblock When will you do what?-{A}nticipating temporal occurrences of
  activities.
\newblock In \emph{IEEE Conference on Computer Vision and Pattern Recognition
  (CVPR)}, 2018.

\bibitem[Abu~Farha et~al.(2020)Abu~Farha, Ke, Schiele, and Gall]{farha2020gcpr}
Y. Abu~Farha, Q. Ke, B. Schiele, and J. Gall.
\newblock Long-term anticipation of activities with cycle consistency.
\newblock In \emph{DAGM German Conference on Pattern Recognition (GCPR)}, 2020.

\bibitem[Ashutosh et~al.(2023)Ashutosh, Girdhar, Torresani, and
  Grauman]{Ashutosh_2023_CVPR}
K. Ashutosh, R. Girdhar, L. Torresani, and K. Grauman.
\newblock Hiervl: Learning hierarchical video-language embeddings.
\newblock In \emph{IEEE Conference on Computer Vision and Pattern Recognition
  (CVPR)}, 2023.

\bibitem[Ba(2016)]{ba2016layer}
Jimmy~Lei Ba.
\newblock Layer normalization.
\newblock \emph{arXiv preprint arXiv:1607.06450}, 2016.

\bibitem[Bahrami et~al.(2023)Bahrami, Francesca, and Gall]{ltc2023bahrami}
Emad Bahrami, Gianpiero Francesca, and Juergen Gall.
\newblock How much temporal long-term context is needed for action
  segmentation?
\newblock In \emph{IEEE International Conference on Computer Vision (ICCV)},
  2023.

\bibitem[Das and Ryoo(2022)]{Das2022VideoC}
S. Das and M.~S. Ryoo.
\newblock Video + clip baseline for ego4d long-term action anticipation.
\newblock \emph{arXiv preprint arXiv:2207.00579}, 2022.

\bibitem[Farha and Gall(2019)]{farha2019uaaa}
Y. Farha and J. Gall.
\newblock Uncertainty-aware anticipation of activities.
\newblock In \emph{IEEE International Conference on Computer Vision Workshop
  (ICCVW)}, 2019.

\bibitem[Furnari and Farinella(2020)]{furnari2020rulstm}
A. Furnari and G.~M. Farinella.
\newblock Rolling-unrolling lstms for action anticipation from first-person
  video.
\newblock \emph{IEEE Transactions on Pattern Analysis and Machine Intelligence
  (TPAMI)}, 2020.

\bibitem[Girase et~al.(2023)Girase, Agarwal, Choi, and
  Mangalam]{harshayu2023latency}
Harshayu Girase, Nakul Agarwal, Chiho Choi, and Karttikeya Mangalam.
\newblock Latency matters: Real-time action forecasting transformer.
\newblock In \emph{IEEE Conference on Computer Vision and Pattern Recognition
  (CVPR)}, 2023.

\bibitem[Girdhar and Grauman(2021)]{girdhar2021anticipative}
R. Girdhar and K. Grauman.
\newblock {Anticipative Video Transformer}.
\newblock In \emph{IEEE International Conference on Computer Vision (ICCV)},
  2021.

\bibitem[Gong et~al.(2022)Gong, Lee, Kim, Ha, and Cho]{gong2022future}
D. Gong, J. Lee, M. Kim, S.J. Ha, and M. Cho.
\newblock Future transformer for long-term action anticipation.
\newblock In \emph{IEEE Conference on Computer Vision and Pattern Recognition
  (CVPR)}, 2022.

\bibitem[Grauman et~al.(2022)Grauman, Westbury, and et~al.]{Grauman2021Ego4DAT}
K. Grauman, A. Westbury, and et al.
\newblock Ego4d: Around the world in 3,000 hours of egocentric video.
\newblock \emph{IEEE Conference on Computer Vision and Pattern Recognition
  (CVPR)}, 2022.

\bibitem[Gu and Dao(2023)]{gu2023mamba}
Albert Gu and Tri Dao.
\newblock Mamba: Linear-time sequence modeling with selective state spaces.
\newblock \emph{arXiv preprint arXiv:2312.00752}, 2023.

\bibitem[Gu et~al.(2021)Gu, Johnson, Goel, Saab, Dao, Rudra, and
  R\'{e}]{NEURIPS2021_05546b0e}
Albert Gu, Isys Johnson, Karan Goel, Khaled Saab, Tri Dao, Atri Rudra, and
  Christopher R\'{e}.
\newblock Combining recurrent, convolutional, and continuous-time models with
  linear state space layers.
\newblock In \emph{Advances in Neural Information Processing Systems
  (NeurIPS)}, 2021.

\bibitem[Gu et~al.(2022)Gu, Goel, and Re]{gu2022efficientlyS4}
Albert Gu, Karan Goel, and Christopher Re.
\newblock Efficiently modeling long sequences with structured state spaces.
\newblock In \emph{International Conference on Learning Representations
  (ICLR)}, 2022.

\bibitem[Guo et~al.(2024)Guo, Agarwal, Lo, Lee, and Ji]{hongji2024unc}
Hongji Guo, Nakul Agarwal, Shao-Yuan Lo, Kwonjoon Lee, and Qiang Ji.
\newblock Uncertainty-aware action decoupling transformer for action
  anticipation.
\newblock In \emph{IEEE Conference on Computer Vision and Pattern Recognition
  (CVPR)}, 2024.

\bibitem[Guo et~al.(2025)Guo, Li, Dai, Ouyang, Ren, and Xia]{guo2025mambair}
Hang Guo, Jinmin Li, Tao Dai, Zhihao Ouyang, Xudong Ren, and Shu-Tao Xia.
\newblock Mambair: A simple baseline for image restoration with state-space
  model.
\newblock In \emph{European Conference on Computer Vision (ECCV)}, 2025.

\bibitem[Han et~al.(2024)Han, Wang, Xia, Han, Pu, Ge, Song, Song, Zheng, and
  Huang]{han2024demystify}
Dongchen Han, Ziyi Wang, Zhuofan Xia, Yizeng Han, Yifan Pu, Chunjiang Ge, Jun
  Song, Shiji Song, Bo Zheng, and Gao Huang.
\newblock Demystify mamba in vision: A linear attention perspective.
\newblock \emph{Advances in Neural Information Processing Systems (NeurIPS)},
  2024.

\bibitem[Hendrycks and Gimpel(2016)]{Hendrycks2016GaussianEL}
Dan Hendrycks and Kevin Gimpel.
\newblock Gaussian error linear units (gelus).
\newblock \emph{arXiv preprint arXiv:1606.08415}, 2016.

\bibitem[Ho et~al.(2020)Ho, Jain, and Abbeel]{ho2020denoisingDiff}
Jonathan Ho, Ajay Jain, and Pieter Abbeel.
\newblock Denoising diffusion probabilistic models.
\newblock \emph{Advances in Neural Information Processing Systems (NeurIPS)},
  2020.

\bibitem[Hu et~al.(2018)Hu, Shen, Albanie, Sun, and
  Wu]{Hu2017SqueezeandExcitationN}
Jie Hu, Li Shen, Samuel Albanie, Gang Sun, and Enhua Wu.
\newblock Squeeze-and-excitation networks.
\newblock \emph{IEEE Conference on Computer Vision and Pattern Recognition
  (CVPR)}, 2018.

\bibitem[Hu et~al.(2024)Hu, Baumann, Gui, Grebenkova, Ma, Fischer, and
  Ommer]{hu2024zigma}
Vincent~Tao Hu, Stefan~Andreas Baumann, Ming Gui, Olga Grebenkova, Pingchuan
  Ma, Johannes Fischer, and Björn Ommer.
\newblock Zigma: A dit-style zigzag mamba diffusion model.
\newblock In \emph{European Conference on Computer Vision (ECCV)}, 2024.

\bibitem[Ke et~al.(2019)Ke, Fritz, and Schiele]{Ke_2019_CVPR}
Q. Ke, M. Fritz, and B. Schiele.
\newblock Time-conditioned action anticipation in one shot.
\newblock In \emph{IEEE Conference on Computer Vision and Pattern Recognition
  (CVPR)}, 2019.

\bibitem[Kingma and Ba(2015)]{kingma2014adam}
Diederik~P. Kingma and Jimmy Ba.
\newblock Adam: {A} method for stochastic optimization.
\newblock In \emph{International Conference on Learning Representations
  (ICLR)}, 2015.

\bibitem[Kuehne et~al.(2014)Kuehne, Arslan, and Serre]{Kuehne12}
Hilde Kuehne, Ali Arslan, and Thomas Serre.
\newblock The language of actions: Recovering the syntax and semantics of
  goal-directed human activities.
\newblock In \emph{IEEE Conference on Computer Vision and Pattern Recognition
  (CVPR)}, 2014.

\bibitem[Li et~al.(2025{\natexlab{a}})Li, Li, Wang, He, Wang, Wang, and
  Qiao]{li2025videomamba}
Kunchang Li, Xinhao Li, Yi Wang, Yinan He, Yali Wang, Limin Wang, and Yu Qiao.
\newblock Videomamba: State space model for efficient video understanding.
\newblock In \emph{European Conference on Computer Vision (ECCV)},
  2025{\natexlab{a}}.

\bibitem[Li et~al.(2025{\natexlab{b}})Li, Singh, and Grover]{li2025mambaND}
Shufan Li, Harkanwar Singh, and Aditya Grover.
\newblock Mamba-nd: Selective state space modeling for multi-dimensional data.
\newblock In \emph{European Conference on Computer Vision (ECCV)},
  2025{\natexlab{b}}.

\bibitem[Liang et~al.(2024)Liang, Zhou, Xu, Zhu, Zou, Ye, Tan, and
  Bai]{liang2024pointmamba}
Dingkang Liang, Xin Zhou, Wei Xu, Xingkui Zhu, Zhikang Zou, Xiaoqing Ye, Xiao
  Tan, and Xiang Bai.
\newblock Pointmamba: A simple state space model for point cloud analysis.
\newblock In \emph{Advances in Neural Information Processing Systems
  (NeurIPS)}, 2024.

\bibitem[Lin et~al.(2020)Lin, Gan, Wang, and Han]{Lin2020TSMTS}
J. Lin, C. Gan, K. Wang, and S. Han.
\newblock Tsm: Temporal shift module for efficient and scalable video
  understanding on edge devices.
\newblock \emph{IEEE Transactions on Pattern Analysis and Machine Intelligence
  (TPAMI)}, 2020.

\bibitem[Liu et~al.(2020)Liu, Tang, Li, and Rehg]{Liu_2020_ECCV}
Miao Liu, Siyu Tang, Yin Li, and James~M Rehg.
\newblock Forecasting human-object interaction: joint prediction of motor
  attention and actions in first person video.
\newblock In \emph{European Conference on Computer Vision (ECCV)}, 2020.

\bibitem[Liu et~al.(2024)Liu, Tian, Zhao, Yu, Xie, Wang, Ye, and
  Liu]{liu2024vmamba}
Yue Liu, Yunjie Tian, Yuzhong Zhao, Hongtian Yu, Lingxi Xie, Yaowei Wang,
  Qixiang Ye, and Yunfan Liu.
\newblock Vmamba: Visual state space model.
\newblock \emph{Advances in Neural Information Processing Systems (NeurIPS)},
  2024.

\bibitem[Liu et~al.(2021)Liu, Lin, Cao, Hu, Wei, Zhang, Lin, and
  Guo]{liu2021Swin}
Ze Liu, Yutong Lin, Yue Cao, Han Hu, Yixuan Wei, Zheng Zhang, Stephen Lin, and
  Baining Guo.
\newblock Swin transformer: Hierarchical vision transformer using shifted
  windows.
\newblock In \emph{IEEE International Conference on Computer Vision (ICCV)},
  2021.

\bibitem[Ma et~al.(2024)Ma, Li, and Wang]{ma2024uMamba}
Jun Ma, Feifei Li, and Bo Wang.
\newblock U-mamba: Enhancing long-range dependency for biomedical image
  segmentation.
\newblock \emph{arXiv preprint arXiv:2401.04722}, 2024.

\bibitem[Mascar\'o et~al.(2023)Mascar\'o, Ahn, and Lee]{Mascaro_2023_WACV}
E.V. Mascar\'o, H. Ahn, and D. Lee.
\newblock Intention-conditioned long-term human egocentric action anticipation.
\newblock In \emph{IEEE Winter Conference on Applications of Computer Vision
  (WACV)}, 2023.

\bibitem[Mittal et~al.(2024)Mittal, Agarwal, Lo, and Lee]{mittal2024can}
Himangi Mittal, Nakul Agarwal, Shao-Yuan Lo, and Kwonjoon Lee.
\newblock Can't make an omelette without breaking some eggs: Plausible action
  anticipation using large video-language models.
\newblock In \emph{IEEE Conference on Computer Vision and Pattern Recognition
  (CVPR)}, 2024.

\bibitem[Mo and Tian(2024)]{mo2024scaling}
Shentong Mo and Yapeng Tian.
\newblock Scaling diffusion mamba with bidirectional ssms for efficient image
  and video generation.
\newblock \emph{arXiv preprint arXiv:2405.15881}, 2024.

\bibitem[Nagarajan et~al.(2020)Nagarajan, Li, Feichtenhofer, and
  Grauman]{ego-topo}
T. Nagarajan, Y. Li, C. Feichtenhofer, and K. Grauman.
\newblock Ego-topo: Environment affordances from egocentric video.
\newblock In \emph{IEEE Conference on Computer Vision and Pattern Recognition
  (CVPR)}, 2020.

\bibitem[Nawhal et~al.(2022)Nawhal, Jyothi, and Mori]{nawhal2022anticipatr}
M. Nawhal, A.~A. Jyothi, and G. Mori.
\newblock Rethinking learning approaches for long-term action anticipation.
\newblock In \emph{European Conference on Computer Vision (ECCV)}, 2022.

\bibitem[Ramachandran et~al.(2017)Ramachandran, Zoph, and
  Le]{Ramachandran2017swish}
Prajit Ramachandran, Barret Zoph, and Quoc~V Le.
\newblock Swish: a self-gated activation function.
\newblock \emph{arXiv preprint arXiv:1710.05941}, 2017.

\bibitem[Ramesh et~al.(2022)Ramesh, Dhariwal, Nichol, Chu, and
  Chen]{ramesh2022hierarchical}
Aditya Ramesh, Prafulla Dhariwal, Alex Nichol, Casey Chu, and Mark Chen.
\newblock Hierarchical text-conditional image generation with clip latents.
\newblock \emph{arXiv preprint arXiv:2204.06125}, 2022.

\bibitem[Roy et~al.(2024)Roy, Rajendiran, and Fernando]{debatiya2024inter}
Debaditya Roy, Ramanathan Rajendiran, and Basura Fernando.
\newblock Interaction region visual transformer for egocentric action
  anticipation.
\newblock In \emph{IEEE/CVF Winter Conference on Applications of Computer
  Vision (WACV)}, 2024.

\bibitem[Ruan and Xiang(2024)]{ruan2024VmUnet}
Jiacheng Ruan and Suncheng Xiang.
\newblock Vm-unet: Vision mamba unet for medical image segmentation.
\newblock \emph{arXiv preprint arXiv:2402.02491}, 2024.

\bibitem[Sener et~al.(2020)Sener, Singhania, and Yao]{sener2020temporal}
Fadime Sener, Dipika Singhania, and Angela Yao.
\newblock Temporal aggregate representations for long-range video
  understanding.
\newblock In \emph{European Conference on Computer Vision (ECCV)}, 2020.

\bibitem[Sener et~al.(2022)Sener, Chatterjee, Shelepov, He, Singhania, Wang,
  and Yao]{sener2022assembly101}
F. Sener, D. Chatterjee, D. Shelepov, K. He, D. Singhania, R. Wang, and A. Yao.
\newblock Assembly101: A large-scale multi-view video dataset for understanding
  procedural activities.
\newblock \emph{IEEE Conference on Computer Vision and Pattern Recognition
  (CVPR)}, 2022.

\bibitem[Shi et~al.(2024)Shi, Dong, and Xu]{shi2024multiscale}
Yuheng Shi, Minjing Dong, and Chang Xu.
\newblock Multi-scale vmamba: Hierarchy in hierarchy visual state space model.
\newblock \emph{arXiv preprint arXiv:2405.14174}, 2024.

\bibitem[Smith et~al.(2023)Smith, Warrington, and
  Linderman]{smith2023simplifiedS6}
Jimmy~T.H. Smith, Andrew Warrington, and Scott Linderman.
\newblock Simplified state space layers for sequence modeling.
\newblock In \emph{International Conference on Learning Representations
  (ICLR)}, 2023.

\bibitem[Sohl-Dickstein et~al.(2015)Sohl-Dickstein, Weiss, Maheswaranathan, and
  Ganguli]{sohl2015NonEquTherm}
Jascha Sohl-Dickstein, Eric Weiss, Niru Maheswaranathan, and Surya Ganguli.
\newblock Deep unsupervised learning using nonequilibrium thermodynamics.
\newblock In \emph{International Conference on Machine Learning (ICML)}, 2015.

\bibitem[Song et~al.(2021{\natexlab{a}})Song, Meng, and
  Ermon]{song2021denoising}
Jiaming Song, Chenlin Meng, and Stefano Ermon.
\newblock Denoising diffusion implicit models.
\newblock In \emph{International Conference on Learning Representations
  (ICLR)}, 2021{\natexlab{a}}.

\bibitem[Song and Ermon(2019)]{song2019generative}
Yang Song and Stefano Ermon.
\newblock Generative modeling by estimating gradients of the data distribution.
\newblock \emph{Advances in Neural Information Processing Systems (NeurIPS)},
  2019.

\bibitem[Song et~al.(2021{\natexlab{b}})Song, Sohl-Dickstein, Kingma, Kumar,
  Ermon, and Poole]{song2021scorebased}
Yang Song, Jascha Sohl-Dickstein, Diederik~P Kingma, Abhishek Kumar, Stefano
  Ermon, and Ben Poole.
\newblock Score-based generative modeling through stochastic differential
  equations.
\newblock In \emph{International Conference on Learning Representations
  (ICLR)}, 2021{\natexlab{b}}.

\bibitem[Stein and McKenna(2013)]{Stein2013CombiningEA}
Sebastian Stein and Stephen~J. McKenna.
\newblock Combining embedded accelerometers with computer vision for
  recognizing food preparation activities.
\newblock \emph{ACM international joint conference on Pervasive and ubiquitous
  computing}, 2013.

\bibitem[Vaswani et~al.(2017)Vaswani, Shazeer, Parmar, Uszkoreit, Jones, Gomez,
  Kaiser, and Polosukhin]{vaswani2017attention}
A. Vaswani, N. Shazeer, N. Parmar, J. Uszkoreit, L. Jones, A.~N. Gomez, L.
  Kaiser, and I. Polosukhin.
\newblock Attention is all you need.
\newblock In \emph{Advances in Neural Information Processing Systems
  (NeurIPS)}, 2017.

\bibitem[Wang et~al.(2023)Wang, Zhu, Wang, Yu, Liu, Omar, and
  Hamid]{wang2023selectiveVideo}
Jue Wang, Wentao Zhu, Pichao Wang, Xiang Yu, Linda Liu, Mohamed Omar, and
  Raffay Hamid.
\newblock Selective structured state-spaces for long-form video understanding.
\newblock In \emph{IEEE Conference on Computer Vision and Pattern Recognition
  (CVPR)}, 2023.

\bibitem[Xing et~al.(2024)Xing, Ye, Yang, Liu, and Zhu]{xing2024segmamba}
Zhaohu Xing, Tian Ye, Yijun Yang, Guang Liu, and Lei Zhu.
\newblock Segmamba: Long-range sequential modeling mamba for 3d medical image
  segmentation.
\newblock In \emph{International Conference on Medical Image Computing and
  Computer-Assisted Intervention}, 2024.

\bibitem[Yan et~al.(2024)Yan, Gu, and Rush]{yan2024diffusion}
Jing~Nathan Yan, Jiatao Gu, and Alexander~M Rush.
\newblock Diffusion models without attention.
\newblock In \emph{IEEE Conference on Computer Vision and Pattern Recognition
  (CVPR)}, pages 8239--8249, 2024.

\bibitem[Zatsarynna and Gall(2023)]{zatsarynna2023goal}
O. Zatsarynna and J. Gall.
\newblock Action anticipation with goal consistency.
\newblock In \emph{IEEE International Conference on Image Processing (ICIP)},
  2023.

\bibitem[Zatsarynna et~al.()Zatsarynna, Bahrami, Farha, Francesca, and
  Gall]{zatsarynna2024gtd}
Olga Zatsarynna, Emad Bahrami, Yazan~Abu Farha, Gianpiero Francesca, and
  Juergen Gall.
\newblock Gated temporal diffusion for stochastic long-term dense anticipation.
\newblock \emph{European Conference on Computer Vision (ECCV)}.

\bibitem[Zatsarynna et~al.(2021)Zatsarynna, Farha, and
  Gall]{zatsarynna2021MMTCN}
O. Zatsarynna, Y. Farha, and J. Gall.
\newblock Multi-modal temporal convolutional network for anticipating actions
  in egocentric videos.
\newblock In \emph{IEEE Conference on Computer Vision and Pattern Recognition
  Workshop (CVPRW)}, 2021.

\bibitem[Zatsarynna et~al.(2022)Zatsarynna, Farha, and
  Gall]{zatsarynna_2022_gcpr}
Olga Zatsarynna, Yazan~Abu Farha, and Juergen Gall.
\newblock Self-supervised learning for unintentional action prediction.
\newblock In \emph{DAGM German Conference on Pattern Recognition (GCPR)}, 2022.

\bibitem[Zhang et~al.(2024)Zhang, Liu, Reid, Hartley, Zhuang, and
  Tang]{zhang2025motion}
Zeyu Zhang, Akide Liu, Ian Reid, Richard Hartley, Bohan Zhuang, and Hao Tang.
\newblock Motion mamba: Efficient and long sequence motion generation.
\newblock In \emph{European Conference on Computer Vision (ECCV)}, 2024.

\bibitem[Zhao and Wildes(2020)]{zhao2020async}
H. Zhao and R.~P. Wildes.
\newblock On diverse asynchronous activity anticipation.
\newblock In \emph{European Conference on Computer Vision (ECCV)}, 2020.

\bibitem[Zhao et~al.(2024)Zhao, Zhang, Wang, Fu, Agarwal, Lee, and
  Sun]{Zhao2023AntGPTCL}
Qi Zhao, Ce Zhang, Shijie Wang, Changcheng Fu, Nakul Agarwal, Kwonjoon Lee, and
  Chen Sun.
\newblock Antgpt: Can large language models help long-term action anticipation
  from videos?
\newblock \emph{Internantional Conference on Learning Representations (ICLR)},
  2024.

\bibitem[Zhao and Kr{\"a}henb{\"u}hl(2022)]{zhao2022testra}
Y. Zhao and P. Kr{\"a}henb{\"u}hl.
\newblock Real-time online video detection with temporal smoothing
  transformers.
\newblock In \emph{European Conference on Computer Vision (ECCV)}, 2022.

\bibitem[Zhong et~al.(2023{\natexlab{a}})Zhong, Schneider, Voit, Stiefelhagen,
  and Beyerer]{Zhong2022AnticipativeFF}
Zeyun Zhong, David Schneider, Michael Voit, Rainer Stiefelhagen, and J{\"u}rgen
  Beyerer.
\newblock Anticipative feature fusion transformer for multi-modal action
  anticipation.
\newblock In \emph{IEEE Winter Conference on Applications of Computer Vision
  (WACV)}, 2023{\natexlab{a}}.

\bibitem[Zhong et~al.(2023{\natexlab{b}})Zhong, Wu, Martin, Voit, Gall, and
  Beyerer]{zhong2023diffant}
Zeyun Zhong, Chengzhi Wu, Manuel Martin, Michael Voit, Juergen Gall, and
  J{\"u}rgen Beyerer.
\newblock Diffant: Diffusion models for action anticipation.
\newblock \emph{arXiv preprint arXiv:2311.15991}, 2023{\natexlab{b}}.

\bibitem[Zhu et~al.(2024)Zhu, Liao, Zhang, Wang, Liu, and
  Wang]{zhu2024visionMamba}
Lianghui Zhu, Bencheng Liao, Qian Zhang, Xinlong Wang, Wenyu Liu, and Xinggang
  Wang.
\newblock Vision mamba: Efficient visual representation learning with
  bidirectional state space model.
\newblock \emph{International Conference on Machine Learning (ICML)}, 2024.

\end{thebibliography}
}


\end{document}